\documentclass[sigconf]{acmart}
\AtBeginDocument{%
  }

\usepackage{amssymb}
\usepackage{pifont}
\usepackage{enumitem}
\usepackage{colortbl}
\setcopyright{acmlicensed}
\usepackage{diagbox}
\usepackage{xcolor}
\usepackage{adjustbox}
\usepackage{makecell}
\usepackage{epsfig}
\usepackage{graphicx}
\usepackage{amsmath}
\usepackage{multirow}
\usepackage{array}
\usepackage{booktabs}
\usepackage{colortbl}
\usepackage{xcolor}
\definecolor{Green}{HTML}{228b22}
\definecolor{RED}{HTML}{e9212c}
\definecolor{BLUE}{HTML}{1181b2}

\AtBeginDocument{%
  }

\setcopyright{acmlicensed}
\copyrightyear{2025}
\acmYear{2025}
\acmDOI{XXXXXXX.XXXXXXX}
\acmConference[ACM MM '25]{Proceedings of the 33rd ACM International Conference on Multimedia}{October 27--31, 2025}{Dublin, Ireland}
\acmISBN{978-1-4503-XXXX-XX/25/10018/06}

\begin{document}

\title{Visual Instance-aware Prompt Tuning}

\author{Xi Xiao}
\authornote{Both authors contributed equally to this research.}
\affiliation{%
  \institution{University of Alabama at Birmingham}
  \city{Birmingham}
  \country{United States}}
\email{xxiao@uab.edu}

\author{Yunbei Zhang}
\authornotemark[1]
\affiliation{%
  \institution{Tulane University}
  \city{New Orleans}
  \country{United States}}
\email{yzhang111@tulane.edu}

\author{Xingjian Li}
\affiliation{%
  \institution{Carnegie Mellon University}
  \city{Pittsburgh}
  \country{United States}}
\email{lixj04@gmail.com}

\author{Tianyang Wang}
\authornote{Project Lead.}
\authornotemark[3]
\affiliation{%
  \institution{University of Alabama at Birmingham}
  \city{Birmingham}
  \country{United States}}
\email{tw2@uab.edu}

\author{Xiao Wang}
\affiliation{%
  \institution{Oak Ridge National Laboratory}
  \city{Oak Ridge}
  \country{United States}}
\email{wangx2@ornl.gov}

\author{Yuxiang Wei}
\affiliation{%
  \institution{Georgia Institute of Technology}
  \city{Atlanta}
  \country{United States}}
\email{weiyuxiang@gatech.edu}

\author{Jihun Hamm}
\affiliation{%
  \institution{Tulane University}
  \city{New Orleans}
  \country{United States}}
\email{jhamm3@tulane.edu}

\author{Min Xu}
\authornote{Corresponding author.}
\affiliation{%
  \institution{Carnegie Mellon University}
  \city{Pittsburgh}
  \country{United States}}
\email{mxu1@cs.cmu.edu}

\renewcommand{\shortauthors}{Xiao et al.}
\acmSubmissionID{838}
\begin{abstract}

Visual Prompt Tuning (VPT) has emerged as a parameter-efficient fine-tuning paradigm for vision transformers, with conventional approaches utilizing dataset-level prompts that remain the same across all input instances. We observe that this strategy results in sub-optimal performance due to high variance in downstream datasets. To address this challenge, we propose Visual Instance-aware  Prompt Tuning (ViaPT), which generates instance-aware prompts based on each individual input and fuses them with dataset-level prompts, leveraging Principal Component Analysis (PCA) to retain important prompting information. Moreover, we reveal that VPT-Deep and VPT-Shallow represent two corner cases based on a conceptual understanding, in which they fail to effectively capture instance-specific information, while random dimension reduction on prompts only yields performance between the two extremes. Instead, ViaPT overcomes these limitations by balancing dataset-level and instance-level knowledge, while reducing the amount of learnable parameters compared to VPT-Deep. Extensive experiments across 34 diverse datasets demonstrate that our method consistently outperforms state-of-the-art baselines, establishing a new paradigm for analyzing and optimizing visual prompts for vision transformers.
\end{abstract}





\maketitle

\section{Introduction}\label{sec:intro}

\vspace{0.5em}
\begin{figure}[t]
  \centering
  \includegraphics[width=0.48\textwidth]{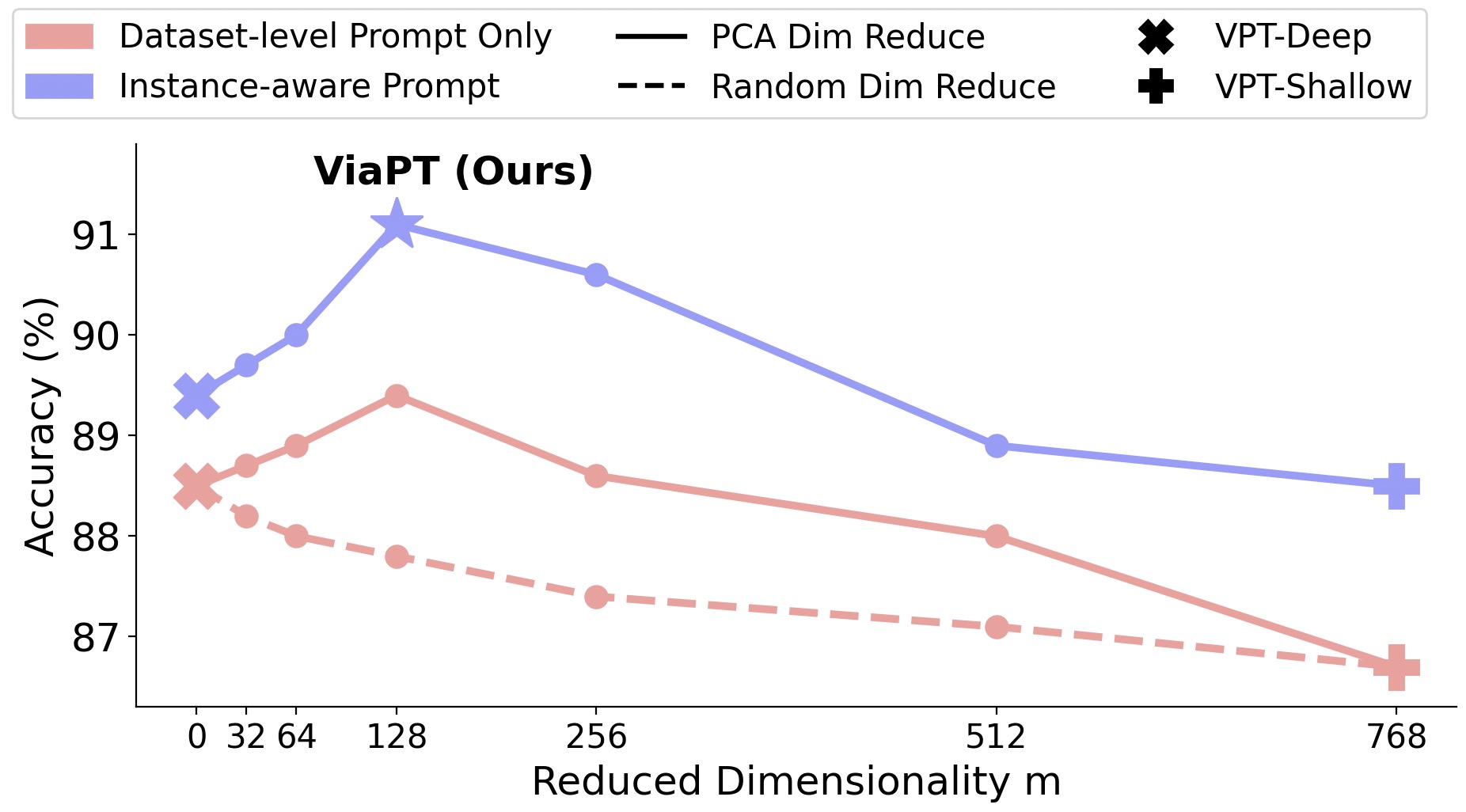}
  \caption{Balancing information flow and instance awareness for optimal adaptation. Our framework (pink) reveals VPT-Shallow and VPT-Deep as corner cases (m=768 and m=0), demonstrating the importance of controlled information flow across layers. Combining with instance-aware prompts (blue) further improves performance across all configurations, with peak accuracy at a medium dimensionality (m=128) highlighting the dual benefit of instance awareness and balanced information propagation.
  }
  \label{fig:intro_curve}
  \vspace{-1em}
\end{figure}

Large-scale vision transformers (ViTs) \cite{dosovitskiy2020image} pretrained on millions of images have become the de-facto backbone for a wide spectrum of visual recognition tasks. Visual Prompt Tuning (VPT) \cite{jia2022visual,bahng2022exploring} has emerged as a parameter-efficient approach for adapting these models to downstream tasks. Instead of updating the backbone through full/partial fine-tuning, a small set of learnable tokens, namely \emph{prompts}, is prepended to the input sequence and optimized while the backbone parameters remain frozen. This strategy has been proven successful in various visual recognition scenarios.

Despite promising, existing VPT methods \cite{jia2022visual,bahng2022exploring,han20232vpt,das2023learning,huang2023hta,pei2024sa2vp,zeng2024visual,jinlor, Zhang_2025_WACV} suffer from a fundamental limitation: the prompt tokens are tied to the entire dataset that is used for tuning the backbone. 
Such dataset-level prompts implicitly assume that a universal set of prompts can adapt the pre-trained backbone to downstream data by reconciling the diverse semantics and visual statistics, overlooking the prompting effect that can be potentially offered by an individual input image (i.e., instance). Moreover, this assumption rarely holds in practice. For example, fine-grained recognition or domain shifts inevitably introduce substantial intra-class variability that cannot be effectively captured by a single set of prompts. Fig.~\ref{fig:intro_curve} illustrates this phenomenon on the CUB-200 dataset: the performance of VPT-Shallow/Deep could be further boosted with instance-aware prompts (pink vs blue). 
Similarly, class imbalance also undermines this assumption, partially explaining why typical VPT methods \cite{jia2022visual,han20232vpt,pei2024sa2vp, Zhang_2025_WACV, zhang2025dpcore} overfit the dominant classes in a dataset. 
These observations suggest that a single set of dataset-level prompts is inherently limited and that instance sensitivity should be encouraged for robust adaptation.

We address this limitation by proposing Visual Instance-aware Prompt Tuning (\textbf{ViaPT}). 
To exploit the prompting effect of an instance, the first step is to determine how to involve the image information in the prompt. A straightforward option could be using a lightweight learnable encoder to extract image features and use them directly as instance-aware prompts. However, we observe that this fashion yields sub-optimal performance, and incurs more learnable parameters.  Therefore, we are motivated to design a new fashion. Specifically, we still use a lightweight encoder that accepts the image as input, aiming to learn the statistics (i.e., mean and std) of the distribution, from which the image comes. Afterwards, we randomly sample variables from $\mathcal{N}(\mathbf{0}, \mathbf{I})$ and then generate instance-aware prompts by scaling and shifting the sampled variables based on the learned mean and std \cite{li2021learning}. To avoid missing global information, we simultaneously use dataset-level prompts along with the proposed instance-aware ones. This is achieved by firstly combining them with regular concatenation, and then applying PCA to the concatenated prompts to retain the most important prompting information. Since VPT requires the same dimension for both image and prompt tokens, we pad the prompts, of which dimension is reduced by PCA, with randomized and learnable parameters, which also increase the dynamics of the prompts. The generation of instance-aware prompts and the use of PCA are integrated into an end-to-end learning pipeline, which has fewer parameters than VPT-Deep  and most of its variants (see Table \ref{tab:full_comparison}). Notably, a certain level of randomness is involved in fine-tuning because of the sampling operations used to generate instance-aware prompts, however, to avoid such randomness during inference, we only sample needed variables once from $\mathcal{N}(\mathbf{0}, \mathbf{I})$ beforehand with a fixed seed for all testing images, thus the instance-aware prompts are mainly determined by the \textit{mean} and \textit{std} associated with a testing image.

Furthermore, in terms of the prompting fashions between the transformer layers in a model, we find that the proposed method along with the typical VPT schemes (i.e., VPT-Deep/Shallow) can be jointly interpreted based on a conceptual understanding. Specifically, 
VPT-Deep, which prepends randomized  prompts to every transformer layer and jointly learns them, corresponds to setting the number of principle components to 0 in PCA (i.e., for the current layer, its prompts are \textit{not} output by the immediate previous layer). In contrast, VPT-Shallow corresponds to setting the number of principle components to the full rank in PCA (i.e., for the current layer, its prompts are output by the immediate previous layer). Random dimension reduction, a naive alternative, falls somewhere in between, but fails to select important components of the prompts,  
while the proposed ViaPT yields a promising balance among these special cases, thereby achieving higher accuracy and stronger generalization. Our extensive experiments across 34 challenging benchmarks demonstrate that ViaPT consistently outperforms state-of-the-art baselines by  on average while still maintaining 
parameter efficiency. 
Moreover, ViaPT is pioneering to shift the conventional dataset-level prompting to instance-level, 
defining a new paradigm for prompt tuning research for vision models. In addition, we also analyze the rationale of combining the instance-aware prompting and PCA in the ablation study in section \ref{sec:ablation}. Notably, while ViaPT coherently integrates both instance-aware prompts generation and PCA, each can still be used with existing VPT schemes as a plug-and-play component. For instance, the yellow curve in Fig. \ref{fig:intro_curve} shows what if only PCA is applied with the conventional VPT, and the two endpoints of the blue curve yield the results of when instance-aware prompting is used with the conventional VPT without PCA.

This work makes the following contributions: (1) we develop a novel visual prompt tuning method that generates instance-aware prompts based on each individual input; (2) we propose to leverage PCA to retain the most important prompting information from both dataset-level and instance-aware prompts; (3) we introduce a conceptual understanding to interpret the proposed method ViaPT along with typical VPT schemes; (4) we demonstrate through extensive experiments on 34 diverse datasets that ViaPT establishes new state of the art while maintaining parameter efficiency.

\section{Related Work}


\subsection{Visual Prompt Tuning}
Visual Prompt Tuning (VPT)~\cite{jia2022visual, bahng2022visual, zeng2024visual, pei2024sa2vp, xiao2025visualvariationalautoencoderprompt,  ren2025vpt, Chowdhury_2025_CVPR, Mai_2025_CVPR, Tu_2023_CVPR, liu2024insvp, ruan2023dynamic, yu2025crisp} has emerged as a powerful parameter-efficient approach for adapting pre-trained Vision Transformers to downstream tasks. Inspired by prompt tuning \cite{lester2021power} in natural language processing, VPT introduces a small set of learnable tokens that are prepended to the input sequence while keeping the backbone model frozen. \cite{jia2022visual,bahng2022exploring} first demonstrated VPT's effectiveness for image classification tasks, showing competitive performance compared to full fine-tuning while updating only 0.1-1\% of parameters. Multi-modal prompt tuning \cite{zhu2023prompt, zhou2022conditional,goswami2024copl, wang2024m2ptmultimodalprompttuning, xiao2025hgtdp} applies this concept to joint vision-language tasks, while prompt ensembling \cite{pitis2023boosted} combines multiple specialized prompts for improved performance. The approach has been extended to test-time adaptation \cite{Zhang_2025_WACV, xiao2024modeladaptationtesttime, niu2024test, zhang2025dpcore, li2024gca}, where prompts are quickly optimized to handle distribution shifts without accessing training data. Despite these advances, existing VPT methods share a critical limitation: they optimize a single set of prompts at the dataset level, which remains static across all input instances. Our work addresses this limitation by introducing instance-specific prompts that dynamically adapt to individual inputs, significantly improving performance on diverse and fine-grained classification tasks.

\subsection{Instance-Aware Adaptation in Vision Models}
The importance of instance-specific adaptations has been recognized across various computer vision tasks \cite{zheng2022promptvisiontransformerdomain, zhou2022cocoop, Oh_2023_CVPR, han2023flatten, xiao2025visualvariationalautoencoderprompt}. Dynamic networks ~\cite{bengio2015conditional, liu2018dynamic, han2023flatten, li2025magicid, yu2025prnet} that adjust their computation paths based on input content have shown improved efficiency and accuracy. In the context of few-shot learning, instance-dependent prototypes \cite{yu-etal-2022-dependency} have demonstrated superior performance over fixed representations. More recently, instance-specific meta networks \cite{zhou2022cocoop, zheng2022promptvisiontransformerdomain} generate weights conditioned on input features. These approaches highlight the benefits of tailoring model behavior to individual inputs. Our work applies this insight to visual prompt tuning by generating instance-specific prompts, 
enabling more flexible adaptation to diverse visual patterns while maintaining parameter efficiency.


\subsection{Information Flow in Deep Neural Networks}
Understanding and optimizing information flow across network layers is crucial for deep learning performance \cite{achtibat2023attribution, huang2017densely, ba2016layer, ji2025cibrcrossmodalinformationbottleneck}. ResNet's skip connections ~\cite{he2016deep} and DenseNet's dense connectivity ~\cite{huang2017densely} demonstrate how direct paths for information propagation can mitigate gradient problems and improve feature reuse. In transformers, techniques like layer normalization ~\cite{ba2016layer} and attention mechanisms facilitate information flow across tokens. More recently, there has been growing interest in selective feature propagation ~\cite{zeng2022not, achtibat2023attribution}, where only the most relevant information is preserved across layers. Our unified prompt framework relates to this line of research by using PCA-based dimension reduction to control information flow between transformer layers, effectively balancing between preserving valuable context and enabling new adaptations. This approach reveals connections between existing VPT variants while establishing a more flexible and effective paradigm for visual prompt tuning.


\section{Methodology}
\label{sec:method}
\begin{figure*}[t]
    \centering
    \includegraphics[width=0.95\linewidth]{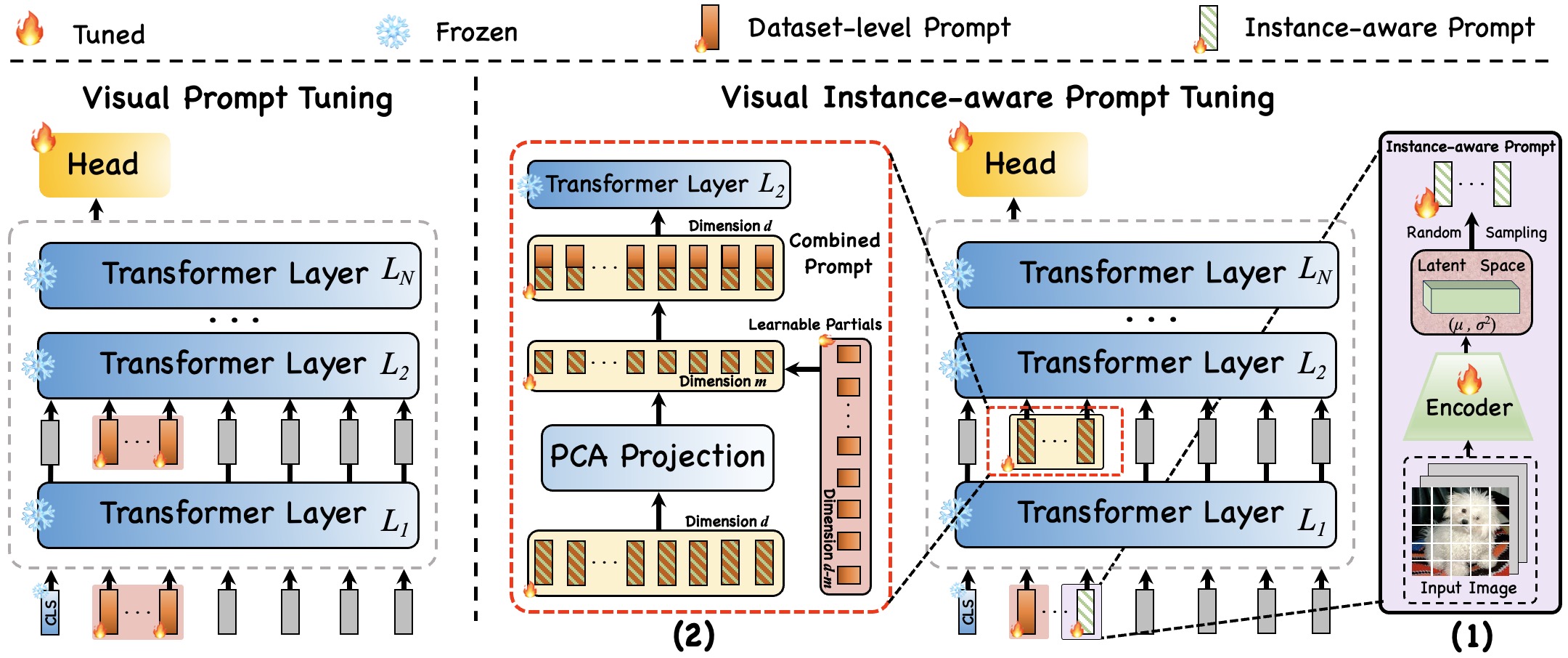}
    \caption{
    \textbf{Overview of the proposed ViaPT framework.}
    Left: Conventional VPT uses dataset-level learnable prompts that remain identical across all inputs.
    Right: Our \textbf{ViaPT} introduces: (1) an Instance-aware Prompt Generator that creates input-specific prompts through a minimal encoder using the reparameterization trick \cite{kingma2013auto}; (2) a Balanced Prompt Propagation module that combines \textcolor{blue}{PCA-projected outputs} (to retain high-variance components in dimension $m$) from previous layers with \textcolor{red}{learnable parameters} (dimension $d-m$) to optimize information flow across transformer layers.
    }
    \label{fig:framework}
    
\end{figure*}

In this section, we present our \textbf{ViaPT} (Visual Instance-aware Prompt Tuning), a framework (Fig. \ref{fig:framework}) that overcomes the limitations of static prompts in visual prompt tuning. After introducing background on conventional VPT (§\ref{subsec:prelim}), we describe our instance-aware prompt generation approach 
(§\ref{subsec:latent-prompt}), present a unified framework for balanced prompt propagation across layers (§\ref{subsec:unified-framework}), and discuss strategies to ensure inference stability (§\ref{subsec:inference}).

\subsection{Preliminaries}
\label{subsec:prelim}
\textbf{Visual Prompt Tuning (VPT).} Vision Transformers (ViT) process input images by splitting them into $k$ patches, which are subsequently transformed into image tokens $\mathbf{E}_0 \in \mathbb{R}^{k \times d}$, where $d$ is the embedding dimension. These image tokens, along with a class token $\mathbf{x}_0$ (often denoted as \texttt{[CLS]}), are processed through $N$ transformer layers $\{L_i\}_{i=1}^N$ to produce the final representation. Visual Prompt Tuning adapts pre-trained ViTs to downstream tasks by introducing a set of $p$ learnable prompt tokens $\mathbf{P} \in \mathbb{R}^{p \times d}$ while keeping the backbone frozen. 

Two primary variants exist: VPT-Shallow and VPT-Deep. In VPT-Shallow, prompts are inserted only at the first transformer layer ($L_1$), resulting in
\begin{align}
\label{eq:vpt_shallow}
    [\mathbf{x}_1,\; \mathbf{Z}_1, \;\mathbf{E}_1] &= \textcolor{blue}{L_1}([\textcolor{blue}{\mathbf{x}_0},\; \textcolor{red}{\mathbf{P}_0},\;\textcolor{blue}{\mathbf{E}_0}] ), \\
    [\mathbf{x}_i,\; \mathbf{Z}_i, \;\mathbf{E}_i] &= \textcolor{blue}{L_i}([\mathbf{x}_{i-1},\; \mathbf{Z}_{i-1},\;\mathbf{E}_{i-1}] ) \quad \text{for } i = 2, \dots, N,\\
    \mathbf{y} &= \textcolor{red}{\mathrm{Head}}(\mathbf{x}_N),
\end{align}
where $\mathbf{Z}_i \in \mathbb{R}^{p \times d}$ represents the features computed by the $i$-th Transformer layer and $[\cdot\;, \;\cdot]$ denotes token concatenation. The colors 
\raisebox{0.1ex}{\textcolor{red}{\rule{0.5em}{0.5em}}} and \raisebox{0.1ex}{\textcolor{blue}{\rule{0.5em}{0.5em}}} indicate \textcolor{red}{trainable} and \textcolor{blue}{frozen} parameters, respectively. In contrast, VPT-Deep introduces unique prompt tokens at each transformer layer's input space: 
\begin{align}
\label{eq:vpt_deep}
    [\mathbf{x}_i,\; \mathbf{Z}_i, \;\mathbf{E}_i] &= \textcolor{blue}{L_i}([\mathbf{x}_{i-1},\; \textcolor{red}{\mathbf{P}_{i-1}},\;\mathbf{E}_{i-1}] ) \quad \text{for } i = 1, \dots, N,\\
    \mathbf{y} &= \textcolor{red}{\mathrm{Head}}(\mathbf{x}_N),
\end{align}
where $\mathbf{P}_{i-1} \in \mathbb{R}^{p \times d}$ are the layer-specific learnable prompts. 

These learnable parameters are optimized over the entire training set and remain fixed during inference, making them effective task-specific prompts. However, this dataset-level optimization often fails to account for instance-specific variations, leading to suboptimal performance when faced with diverse visual patterns within the same task.

\subsection{Instance-aware Prompt Generation}
\label{subsec:latent-prompt}

To move beyond dataset-level static prompts and account for input-specific visual variations, we divide the original prompts in the first layer $\mathbf{P}_0:=\{\mathbf{p}_0^1, \mathbf{p}_0^2, ..., \mathbf{p}_0^p\}$ into two sets: the first $\lambda$ tokens form the instance-aware prompt $\mathbf{P}_0^{\mathrm{ins}}:=\{\mathbf{p}_0^{1}, \mathbf{p}_0^{2}, ..., \mathbf{p}_0^{\lambda}\}$ and the remaining $p-\lambda$ tokens comprise the dataset-level prompt $\mathbf{P}_0^{\mathrm{dom}}:=\{\mathbf{p}_0^{\lambda+1}, \mathbf{p}_0^{\lambda + 2}, ..., \mathbf{p}_0^{p}\}$. 
The dataset-level prompts are initialized and updated following standard VPT-Shallow/Deep approaches. For instance-aware prompts, our goal is to design a prompt generator that takes input data and produces corresponding prompt tokens.

While using raw images as input is intuitive, it introduces significant computational overhead. Instead, we leverage the image tokens $\mathbf{E}_0$, which represent images in a much lower dimension to improve efficiency. We define the instance prompt generator as $g: \mathbb{R}^{k\times d} \rightarrow \mathbb{R}^{\lambda\times d}$, where for each input $\mathbf{E}_0$, the generator synthesizes $\lambda$ prompt tokens $g(\mathbf{E}_0)$, which are then concatenated with the dataset-level prompts for prompt tuning.

Although the generator introduces additional trainable parameters beyond the prompts, this is necessary to capture instance-level 
characteristics. 
To maintain both efficacy and parameter efficiency, we simply use a learnable 2-layer convolutional encoder
as the generator, which performs well as demonstrated in our experiments. To further reduce trainable parameters in line with PEFT principles, 
we find that learning a probabilistic generator 
is more effective than a deterministic one. This approach reduces the output of $g$ to $\mathbb{R}^{2\times d}$ instead of $\mathbb{R}^{\lambda \times d}$, significantly decreasing parameter count and computation. Furthermore, this operation constrains the instance-aware prompt distribution to a Gaussian $\mathcal{N}(\mathbf{0}, \mathbf{I})$, which aligns with the initialization of dataset-level prompts and mitigates magnitude discrepancies between the two prompt types. For each set of image tokens $\mathbf{E}_0$, the generator $g$ returns mean and standard deviation pairs: $(\mathbf{\mu}, \mathbf{\sigma}) = g(\mathbf{E}_0)$ where $\mathbf{\mu} , \mathbf{\sigma}\in \mathbb{R}^d$. We then sample $\lambda$ times from $\mathcal{N}(\mathbf{0}, \mathbf{I})$ to obtain $\{\mathbf{z}^1, ... \mathbf{z}^\lambda \}$ and transform them with:
\begin{align}
    (\mathbf{\mu}, \mathbf{\sigma}) &= \textcolor{red}{g}(\mathbf{E}_0) \\
    \{\mathbf{z}^1, ... \mathbf{z}^\lambda \} &\sim \mathcal{N}(\mathbf{0}, \mathbf{I}) \\
    \mathbf{p}_0^i &=\mathbf{z}^i \cdot \mathbf{\sigma} + \mathbf{\mu} \quad \mathrm{for }\; i=1, .., \lambda
\end{align}

We employ an additional KL divergence regularization term 
to constrain the instance-aware prompt distribution as mentioned above. 
The parameters of \textcolor{red}{$g$, $\mathbf{P}$} and \textcolor{red}{Head} are optimized using the objective function:
\begin{equation}
    \mathcal{L} = \mathcal{L}_{\mathrm{xent}} + \beta \cdot \mathrm{D}_{\mathrm{KL}}(\mathcal{N}(\mathbf{\mu}, \mathbf{\sigma}^2) \| \mathcal{N}(\mathbf{0}, \mathbf{I}) )
\end{equation}
where $\mathcal{L}_{\mathrm{xent}}$ is the conventional cross-entropy loss and $\mathrm{D}_{\mathrm{KL}}(\cdot, \cdot)$ is the KL divergence.

\subsection{A Unified Framework for Balanced Prompt Propagation}
\label{subsec:unified-framework}

\textbf{Motivations.} Existing VPT variants represent opposite extremes in handling prompt information flow across transformer layers. VPT-Shallow propagates all prompt features from previous layers but lacks layer-specific adaptability, while VPT-Deep discards all previous information and introduces entirely new prompts at each layer, increasing parameter count substantially. As shown in Fig.~\ref{fig:intro_curve}, both approaches are suboptimal: discarding all prompt outputs from previous layers potentially loses valuable instance-related information, while preserving everything leaves no room for layer-specific adaptations. This observation motivates us to develop a balanced framework that selectively retains the most informative components from previous layers while allowing for new learnable parameters.

To achieve this balance, we introduce a dimension reduction strategy based on Principal Component Analysis (PCA). Let $\mathbf{Z}_{i-1} \in \mathbb{R}^{p \times d}$ represent the output prompt features from layer $i-1$. Instead of directly propagating these features (VPT-Shallow) or completely replacing them with new prompt tokens (VPT-Deep), we apply PCA to each prompt token to reduce its dimensionality. Specifically, we define a PCA transformation $\Phi: \mathbb{R}^{p \times d} \rightarrow \mathbb{R}^{p \times m}$ where $m < d$ is the reduced dimension of each prompt token:
\begin{align}
\label{eq:pca-transform}
\mathbf{Z}_{i-1}^{\text{PCA}} = \Phi(\mathbf{Z}_{i-1}) \in \mathbb{R}^{p \times m}
\end{align}
Since the transformer requires input prompts of dimension $d$, we supplement each reduced prompt token with $(d-m)$ learnable parameters. For each transformer layer $i > 1$, we construct the input prompt tokens by concatenating the PCA-reduced components with learnable parameters along the feature dimension:
\begin{align}
\label{eq:combined-prompt}
\textcolor{Green}{\mathbf{P}_{i-1}^{\text{combined}}} = [\textcolor{blue}{\mathbf{Z}_{i-1}^{\text{PCA}}}, \textcolor{red}{\mathbf{P}_{i-1}^{\text{new}}}] \in \mathbb{R}^{p \times d}
\end{align}
where $\mathbf{P}_{i-1}^{\text{new}} \in \mathbb{R}^{p \times (d-m)}$ represents randomly initialized, learnable parameters. The color
\raisebox{0.1ex}{\textcolor{Green}{\rule{0.5em}{0.5em}}} indicates \textcolor{Green}{combined} parameters. In other words, each prompt token in $\textcolor{Green}{\mathbf{P}_{i-1}^{\text{combined}}}$ consists of two parts: the output of PCA in the first $m$ dimensions and continuously learnable parameters in the remaining $d-m$ dimensions. This combined prompt is then fed into the $i$-th transformer layer:
\begin{align}
\label{eq:layer-with-pca}
[\mathbf{x}_i, \mathbf{Z}_i, \mathbf{E}_i] = \textcolor{blue}{L_i}([\mathbf{x}_{i-1}, \textcolor{Green}{\mathbf{P}_{i-1}^{\text{combined}}}, \mathbf{E}_{i-1}])
\end{align}
Our approach provides a continuous spectrum between VPT-Shallow and VPT-Deep through the parameter $m$, which controls how much information from previous layers is preserved within each prompt token. When $m = 0$, our method reduces to VPT-Deep, as no information is carried forward from previous layers and all prompt dimensions are newly learned at each layer. Conversely, when $m = d$, our method becomes equivalent to VPT-Shallow, as all dimensions of prompt tokens from the previous layer are preserved with no new learnable parameters.

By setting $0 < m < d$, we achieve a balanced trade-off that leverages both the global context captured in earlier layers and the layer-specific adaptations. Unlike random dimension reduction, which may discard important features arbitrarily, PCA ensures that we retain the dimensions with the highest variance, effectively capturing the most informative aspects of the prompt representations. This controlled information flow allows ViaPT to maintain parameter efficiency while approaching or exceeding the performance of more parameter-intensive approaches.

The optimal value of $m$ depends on the specific task and dataset characteristics, but our empirical results (Section~\ref{sec:exp}) show that a moderate reduction (typically $m \approx 128$) consistently yields the best performance across diverse benchmarks. This supports our hypothesis that both preserving key information from earlier layers and allowing for layer-specific adaptations are complementary and essential for effective visual prompt tuning.

\subsection{Inference Strategies and Stability}
\label{subsec:inference}

After training our ViaPT model, the inference process presents unique considerations due to the probabilistic nature of the instance-aware prompt generation. The standard inference procedure follows the same pipeline as training.

However, sampling from a Gaussian distribution introduces randomness that can lead to variations in prediction results for the same input across different inference runs. To address this challenge, we investigate several strategies to achieve stable and reliable predictions during inference.

\textbf{Multi-round inference.} Our primary approach involves performing multiple rounds of inference for each input image and aggregating the results. Specifically, for an input image, we conduct $R$ independent forward passes, each with a different set of randomly sampled prompt tokens. The final prediction is obtained by averaging the class probabilities across all rounds: $\mathbf{p}_{\text{final}} = \frac{1}{R}\sum_{r=1}^{R} \mathbf{p}_r$
where $\mathbf{p}_r$ represents the class probability distribution from the $r$-th round. In our experiments, we found that $R=5$ provides a good balance between inference stability and computational efficiency, with the standard deviation of accuracy dropping below 0.1\% across different runs.

\textbf{Deterministic alternatives.} While multi-round inference effectively mitigates the randomness issue, there are scenarios where deterministic predictions are preferred, especially in latency-sensitive applications. We explore two alternatives to achieve deterministic inference:

1. \textit{Direct prompt generation:} Instead of generating distributional parameters $(\mathbf{\mu}, \mathbf{\sigma})$, we can modify the generator $g$ to directly output $\lambda$ prompt tokens: $g': \mathbb{R}^{k \times d} \rightarrow \mathbb{R}^{\lambda \times d}$. This approach eliminates the sampling step altogether, ensuring deterministic predictions at the cost of increased parameter count. Our experiments show that this approach achieves comparable accuracy to the probabilistic version while providing consistent predictions.

2. \textit{Fixed sampling:} After training the model with sampling, we can generate a set of fixed random samples $\{\mathbf{z}^1, ... \mathbf{z}^\lambda \}$ from $\mathcal{N}(\mathbf{0}, \mathbf{I})$ and use these same samples for all images during inference. This maintains the benefits of reduced parameters 
while ensuring deterministic predictions. Surprisingly, this simple approach yields stable performance comparable to multi-round inference, suggesting that the learned $(\mathbf{\mu}, \mathbf{\sigma})$ capture most of the instance-specific information needed for accurate classification.

While both deterministic alternatives offer practical advantages for deployment, we find that the original 
multi-round inference provides the best overall performance across our benchmarks. This suggests that the stochasticity introduced by sampling may actually benefit the model by providing a form of test-time ensemble, similar to the benefits observed in dropout during inference. The choice among these strategies ultimately depends on the specific application requirements, with trade-offs between accuracy, stability, and computational efficiency.

\section{Experiments}
\label{sec:exp}

We conduct extensive experiments to validate the effectiveness and generalization capability of our proposed \textbf{ViaPT} framework across various image classification benchmarks. We begin by comparing ViaPT with state-of-the-art parameter-efficient tuning methods under the ViT-Base backbone to demonstrate overall superiority in both fine-grained and diverse-task scenarios. We further investigate its generalizability by applying the same design to Swin Transformers, revealing that our improvements hold across different vision backbones. To better understand the robustness of ViaPT under different initialization strategies, we evaluate its performance under both supervised and self-supervised pretraining paradigms. In addition, we perform thorough ablation studies to isolate the contribution of each core component, including instance-aware prompts and PCA-based prompt propagation. Finally, we provide qualitative visualizations of the learned representations through t-SNE, Grad-CAM, and token similarity heatmaps, offering deeper insight into the interpretability and internal behaviors of our method. All experiments are conducted under a fixed tuning budget to ensure fair comparison.

\subsection{Experiment Setup}
\label{subsec:exp_setup}
\noindent\textbf{Datasets.} We conduct extensive evaluations on three widely adopted image classification benchmarks: \textbf{HTA}~\cite{huang2023hta}, \textbf{VTAB-1k}~\cite{zhai2019large}, and \textbf{FGVC}~\cite{wah2011caltech}. HTA is a recent benchmark to evaluate prompt tuning methods under diverse task and domain shifts. It comprises ten classification datasets that vary in granularity, modality, and semantic complexity, including CIFAR-10, CIFAR-100~\cite{krizhevsky2009cifar}, DTD~\cite{cimpoi2014describing}, CUB-200~\cite{wah2011cub}, NABirds~\cite{van2015nabirds}, Oxford Flowers~\cite{nilsback2008oxford}, Food101~\cite{bossard2014food}, GTSRB~\cite{stallkamp2012man}, and SVHN~\cite{netzer2011reading}. We follow the official experimental settings provided in DAM-VP~\cite{huang2023hta} to ensure direct comparability with previous works.
VTAB-1k consists of 19 diverse tasks categorized into three groups: (1) \textit{Natural}, involving real-world images captured by standard cameras; (2) \textit{Specialized}, featuring data from domain-specific devices (e.g., satellites, medical scanners); and (3) \textit{Structured}, focusing on geometric reasoning and synthetic scenes (e.g., object counting, position encoding). The structured tasks are semantically distant from the ImageNet-21k~\cite{deng2009imagenet} pretraining distribution, making them more challenging. Each VTAB task includes 1,000 training examples with a standard 800/200 \texttt{train/val} split. The FGVC benchmark covers 5 fine-grained classification datasets: CUB-200-2011, NABirds, Oxford Flowers, Stanford Dogs, and Stanford Cars, where each dataset is split into 90\% \texttt{train} and 10\% \texttt{val}.

\vspace{0.5em} \noindent\textbf{Baselines.} We benchmark ViaPT against a variety of PEFT methods as outlined by \cite{zeng2024visual}. All comparisons utilize ViT-Base/16~\cite{dosovitskiy2020image} backbone pretrained on ImageNet-21k to ensure fairness and reproducibility. Detailed comparisons including both supervised and self-supervised pretrained variants, such as MAE~\cite{he2022masked} and MoCo v3~\cite{chen2021empirical}, are discussed in \S\ref{sec:pretrain}.

\vspace{0.5em} \noindent\textbf{Implementation Details.} We employ the AdamW optimizer, using a cosine decay schedule for the learning rate. Learning rates are selected via a grid search on the validation set from the following values: $\{50, 25, 10, 5, 2.5, 1, 0.5, 0.25, 0.1, 0.05\}$, and weight decay rates from $\{0.01, 0.001, 0.0001, 0.0\}$. All models are trained for 100 epochs. Notably, ViaPT requires no task-specific tuning or unusually high learning rates, maintaining a parameter footprint of only 0.25\% of the total ViT backbone. All experiments are implemented using PyTorch\cite{NEURIPS2019_9015} and conducted on NVIDIA A100-80GB GPUs.

\begin{table*}[!htbp]
\centering
    \caption{
    \textbf{Comparison of fine-tuning methods with ViT-Base/16 backbone.} \textbf{Bold} and \underline{underlined} indicate best and second-best results. We report mean accuracy across datasets for each benchmark (FGVC: 5 datasets, HTA: 10 datasets, VTAB-1k: 19 datasets) and for each VTAB-1k category (Natural, Specialized, and Structured). See Appendix \ref{sec: A} for per-task results.
    Same for Table \ref{table:mae_moco}.}
\label{tab:full_comparison}

\scriptsize 

\begin{adjustbox}{width=0.9\textwidth,center}
\begin{tabular}{c||c|c|c|c|ccc|c} 
\Xhline{4\arrayrulewidth}
\rowcolor{gray!20}
ViT-Base/16~\cite{dosovitskiy2020image} & Tuned/Total & Extra Params & FGVC Mean & HTA Mean & \multicolumn{3}{c|}{VTAB-1k~\cite{zhai2019large}} & VTAB-1k Mean \\ 
\rowcolor{gray!20}
\rowcolor{gray!20}
(85.8M) & (\%) & & (\%) & (\%) & \textit{Natural} & \textit{Specialized} & \textit{Structured} & (\%) \\ 
\hline \hline
Full \textcolor{lightgray}{\scriptsize{[CVPR22]}}\cite{iofinova2022well} & 100.00 & — & 88.54 & 85.8 & 75.88 & 83.36 & 47.64 & 65.57 \\ 
\hline
Linear \textcolor{lightgray}{\scriptsize{[CVPR22]}}\cite{iofinova2022well} & 0.08 & — & 79.32 & 75.7 & 68.93 & 77.16 & 26.84 & 52.94 \\ 
Partial-1 \textcolor{lightgray}{\scriptsize{[NeurIPS14]}}\cite{yosinski2014transferable} & 8.34 & — & 82.63 & 80.8 & 69.44 & 78.53 & 34.17 & 56.52 \\ 
MLP-3 \textcolor{lightgray}{\scriptsize{[CVPR20]}}\cite{chen2020improved} & 1.44 & \checkmark & 79.80 & 78.5 & 67.80 & 72.83 & 30.62 & 53.21 \\ 
\hline
Sidetune \textcolor{lightgray}{\scriptsize{[ECCV20]}}\cite{zhang2020side} & 10.08 & — & 78.35 & 72.3 & 58.21 & 68.12 & 23.41 & 45.65 \\ 
Bias \textcolor{lightgray}{\scriptsize{[NeurIPS17]}}\cite{rebuffi2017learning} & 0.80 & — & 88.41 & 82.1 & 73.30 & 78.25 & 44.09 & 62.05 \\ 
Adapter \textcolor{lightgray}{\scriptsize{[NeurIPS20]}}\cite{cai2020tinytl} & 1.02 & \checkmark & 85.46 & 80.6 & 70.67 & 77.80 & 33.09 & 62.41 \\ 
LoRA \textcolor{lightgray}{\scriptsize{[ICLR22]}}\cite{hu2022lora} & — & \checkmark & 89.46 & 85.5 & 78.26 & 83.78 & 56.20 & 72.25 \\ 
AdaptFormer \textcolor{lightgray}{\scriptsize{[NeurIPS22]}}\cite{chen2022adaptformer} & — & \checkmark & 89.12 & 89.0 & 80.56 & 84.88 & 58.38 & 72.32 \\ 
ARC$_{\text{att}}$ \textcolor{lightgray}{\scriptsize{[NeurIPS23]}}\cite{dong2023efficient} & — & \checkmark & 89.12 & 89.0 & 80.41 & 85.55 & 58.38 & 72.32 \\ 
\hline
VPT-S \textcolor{lightgray}{\scriptsize{[ECCV22]}}\cite{jia2022visual} & 0.16 & \checkmark & 84.62 & 85.5 & 76.81 & 79.66 & 46.98 & 64.85 \\ 
VPT-D \textcolor{lightgray}{\scriptsize{[ECCV22]}}\cite{jia2022visual} & 0.73 & \checkmark & 89.11 & 85.5 & 78.48 & 82.43 & 54.98 & 69.43 \\
E2VPT \textcolor{lightgray}{\scriptsize{[ICCV23]}}\cite{han2023e2vpt} & 0.39 & \checkmark & 89.22 & 88.5 & 80.01 & 84.43 & 57.39 & 71.42 \\ 
EXPRES \textcolor{lightgray}{\scriptsize{[CVPR23]}}\cite{das2023learning} & — & \checkmark & — & — & 79.69 & 84.03 & 54.99 & 70.02 \\ 
DAM-VP \textcolor{lightgray}{\scriptsize{[CVPR23]}}\cite{huang2023hta} & — & \checkmark & — & 88.50 & — & — & — & — \\ 
SA\textsuperscript{2}VP \textcolor{lightgray}{\scriptsize{[AAAI24]}}\cite{pei2024sa2vp} & 0.76 & \checkmark & \underline{90.08} & \underline{91.50} & \underline{80.97} & \textbf{85.73} & \underline{60.80} & \underline{75.83} \\ 
VFPT \textcolor{lightgray}{\scriptsize{[NeurIPS24]}}\cite{zeng2024visual} & 0.66 & \checkmark & 89.24 & — & 81.35 & 84.93 & 60.19 & 75.49 \\
LoR-VP \textcolor{lightgray}{\scriptsize{[ICLR25]}}\cite{jinlor} & - & \checkmark & 89.32 & — & 79.91 & 83.16 & 60.01 & 74.36 \\ 
\hline 
\rowcolor{gray!20} 
\textbf{Ours} & 0.66 & \checkmark & \textbf{91.40} & \textbf{92.20} & \textbf{82.62} & \underline{85.22} & \textbf{61.25} & \textbf{76.36} \\ 
\Xhline{4\arrayrulewidth}
\end{tabular}
\end{adjustbox}
\label{tab:vit}
\end{table*}

\subsection{Comparison with State-of-the-Art Methods}
\label{sec:overall_comparison}

As shown in Table~\ref{tab:vit}, our method consistently achieves the best performance across all evaluated benchmarks. Specifically, we obtain the highest average accuracy on \textbf{FGVC (91.40\%)}, \textbf{HTA (92.20\%)}, and \textbf{VTAB-1k (76.36\%)}, outperforming not only full fine-tuning (88.54\%, 85.80\%, 65.57\%) but also strong prompting baselines such as SA$^2$VP~\cite{pei2024sa2vp} and VFPT~\cite{zeng2024visual}. Particularly on \textit{Structured} tasks in VTAB-1k, which are known to exhibit significant distributional shift, our method achieves 61.25\%, surpassing SA$^2$VP by +0.45\%. It is worth noting that although we retain the same number of prompt tokens as standard VPT, we introduce a semantic bottleneck within each token by projecting into a task-relevant subspace via PCA. This design reorganizes each prompt token’s representation rather than compressing the total token count, effectively filtering out redundant directions while preserving the most informative semantic axes. Compared to VPT-D~\cite{jia2022visual}, which uses a universal prompt shared across all samples, our instance-aware mechanism dynamically generates personalized prompts for each input, significantly improving performance on datasets with large intra-class variation, such as CUB-200, Pets, and DTD. To further disentangle the impact of each component, we provide an ablation study 
in Table~\ref{table:ablation_main}, clearly showing that both instance-aware prompting and PCA projection in our method are essential to the overall improvement. Despite operating under a strict parameter budget (\textbf{0.74\%} trainable parameters), our approach establishes new state-of-the-art results across all settings. Compared to SA$^2$VP~\cite{pei2024sa2vp}, which adopts a similar tuning ratio (0.76\%) but falls short by \textbf{0.85\%} on average, ours achieves better adaptation by jointly leveraging prompt subspace modeling and instance-level 
semantics within a unified framework. In addition, an interesting observation on VTAB-1k is that VPT related methods including ours show better results when source and target domains have a relatively big gap, well aligned with the finding in \cite{han2024facing}.

\noindent
\textbf{Parameter efficiency.} As shown in Table \ref{tab:full_comparison}, our method has fewer learnable parameters than most of the competitors. Notably, while VPT-S (shallow) holds much fewer parameters, it yields much lower performance since randomized learnable parameters are only applied at the input layer. E2VPT focuses on distillation, thus showing an advantage of parameter amount.

\begin{table*}[!ht]
\caption{\textbf{Comparison of the fine-tuning methods under different pretraining paradigms.} 
We report VTAB-1k~\cite{zhai2019large} accuracy using ViT-Base~\cite{dosovitskiy2020image} as the frozen backbone, pretrained with MAE~\cite{he2022masked} and MoCo v3~\cite{chen2021empirical} respectively. 
Our method shows competitive performance with a moderate parameter overhead.
}
\label{table:mae_moco}

\begin{adjustbox}{width=0.75\textwidth,center}
\begin{tabular}{c||r|rrr||r|rrr} 
\Xhline{4\arrayrulewidth}
\rowcolor{gray!20}
\textbf{Pretraining} & \multicolumn{4}{c||}{MAE~\cite{he2022masked}} & \multicolumn{4}{c}{MoCo v3~\cite{chen2021empirical}} \\
\rowcolor{gray!20}
Method & Tuned/\% & Natural & Specialized & Structured & Tuned/\% & Natural & Specialized & Structured \\
\hline \hline
Full \textcolor{lightgray}{\scriptsize{[CVPR22]}}\cite{iofinova2022well} & 100.00\% & 59.31\% & 79.68\% & 53.82\% & 100.00\% & 71.95\% & 84.72\% & 51.98\% \\
\hline
Linear \textcolor{lightgray}{\scriptsize{[CVPR22]}}\cite{iofinova2022well} & 0.04\% & 18.87\% & 53.72\% & 23.70\% & 0.04\% & 67.46\% & 81.08\% & 30.33\% \\
Partial-1 \textcolor{lightgray}{\scriptsize{[NeurIPS14]}}\cite{yosinski2014transferable} & 8.30\% & \textbf{58.44\%} & \textbf{78.28\%} & \underline{47.64\%} & 8.30\% & 72.31\% & 84.58\% & 47.89\% \\
\hline
Bias \textcolor{lightgray}{\scriptsize{[NeurIPS17]}}\cite{rebuffi2017learning} & 0.16\% & 54.55\% & 75.68\% & \textbf{47.70\%} & 0.16\% & 72.89\% & 81.14\% & 53.43\% \\
Adapter \textcolor{lightgray}{\scriptsize{[NeurIPS20]}}\cite{cai2020tinytl} & 0.87\% & \underline{54.90\%} & 75.19\% & 38.98\% & 1.12\% & 74.19\% & 82.66\% & 47.69\% \\
\hline
VPT-S \textcolor{lightgray}{\scriptsize{[ECCV22]}}\cite{jia2022visual} & 0.05\% & 39.96\% & 69.65\% & 27.50\% & 0.06\% & 67.34\% & 82.26\% & 37.55\% \\
VPT-D \textcolor{lightgray}{\scriptsize{[ECCV22]}}\cite{jia2022visual} & 0.31\% & 36.02\% & 60.61\% & 26.57\% & 0.22\% & 70.27\% & 83.04\% & 42.38\% \\
GPT \textcolor{lightgray}{\scriptsize{[arXiv24]}}\cite{yoo2023improving} & 0.05\% & 47.61\% & 76.86\% & 36.80\% & 0.06\% & 74.84\% & 83.38\% & 49.10\% \\
VFPT \textcolor{lightgray}{\scriptsize{[NeurIPS24]}}\cite{zeng2024visual} & 0.38\% & 53.59\% & 77.75\% & 36.15\% & 0.22\% & \underline{77.47\%} & \underline{85.76\%} & \underline{58.74\%} \\
\rowcolor{gray!20}
\textbf{Ours} & 0.36\% & 54.26\% & \underline{78.01\%} & 37.52\% & 0.30\% & \textbf{79.12\%} & \textbf{86.81\%} & \textbf{60.05\%} \\
\hline
\end{tabular}
\end{adjustbox}


\end{table*}

\begin{table}[!htbp]
\centering
\caption{\textbf{Method comparison on VTAB-1k using Swin-Base model pretrained on ImageNet-21k.}}
\label{tab:performance_vtab1k}
\vspace{-3pt}

\scriptsize 
\setlength{\tabcolsep}{3.5pt} 

\begin{adjustbox}{width=0.48\textwidth,center} 
\begin{tabular}{c||c|ccc} 
\Xhline{4\arrayrulewidth}

Swin-Base & Tuned/Total & \multicolumn{3}{c}{VTAB-1k} \\ 

(86.7M) & (\%) & \textit{Natural} & \textit{Specialized} & \textit{Structured} \\ 
\hline \hline
Full \textcolor{lightgray}{\scriptsize{[CVPR22]}}\cite{iofinova2022well} & 100.00 & 79.10 & 86.21 & 59.65 \\ 
\hline
Linear \textcolor{lightgray}{\scriptsize{[CVPR22]}}\cite{iofinova2022well} & 0.06 & 73.52 & 80.77 & 33.52 \\ 
Bias \textcolor{lightgray}{\scriptsize{[NeurIPS17]}}\cite{rebuffi2017learning} & 0.30 & 76.78 & 83.33 & 51.85 \\ 
VPT-deep \textcolor{lightgray}{\scriptsize{[ECCV22]}}\cite{jia2022visual} & 0.25 & 76.78 & 83.33 & 51.85 \\
E\textsuperscript{2}VPT \textcolor{lightgray}{\scriptsize{[ICCV23]}}\cite{han2023e2vpt} & 0.21 & 83.31 & 84.95 & 57.35 \\ 
SA\textsuperscript{2}VP \textcolor{lightgray}{\scriptsize{[AAAI24]}}\cite{pei2024sa2vp} & 0.29 & 80.81 & \textbf{86.30} & \underline{60.03} \\ 
VFPT \textcolor{lightgray}{\scriptsize{[NeurIPS24]}}\cite{zeng2024visual} & 0.27 & \underline{84.53} & 86.15 & 58.21 \\ 
\hline
\rowcolor{gray!20}
\textbf{Ours} & 0.27 & \textbf{85.29} & \underline{86.21} & \textbf{62.24} \\ 
\Xhline{4\arrayrulewidth}
\end{tabular}
\end{adjustbox}

\vspace{-1em}
\end{table}

\subsection{Generalization to Other Backbones}
\label{sec:backbone}

To assess the generality of our method across architectures, we evaluate its performance using the Swin-Base~\cite{liu2021swin} backbone pretrained on ImageNet-21k. Table~\ref{tab:performance_vtab1k} presents the results on VTAB-1k, comparing our approach with both classical and recent parameter-efficient tuning methods.

\vspace{0.3em}
\noindent\textbf{Consistent improvements across all VTAB categories.} Our method achieves the highest accuracy in the \textit{Natural} (85.29\%) and \textit{Structured} (62.24\%) categories, while matching the best in \textit{Specialized} (86.21\%). In particular, our structured-task performance surpasses the strongest baseline (SA\textsuperscript{2}VP) by +2.21\%, suggesting that our method is better at capturing spatial dependencies and relational patterns that are prevalent in counting and geometric reasoning tasks.

\vspace{0.3em}
\noindent\textbf{Improved robustness under architectural shift.}  
Although most visual prompting methods are initially developed for standard ViT architectures, our method retains strong performance when applied to hierarchical Transformer-based backbones such as Swin. This demonstrates that our semantic compression and instance-aware prompting mechanisms are not tightly coupled to a specific ViT formulation, and can effectively generalize across a variety of Transformer-based vision models.

\subsection{Effectiveness Across Pretraining Paradigms}
\label{sec:pretrain}

To evaluate the robustness and generality of our approach under different pretraining paradigms, we compare the methods using ViT-Base/16 models pretrained with either MAE~\cite{he2022masked} or MoCo v3~\cite{chen2021empirical}. As shown in Table~\ref{table:mae_moco}, our method shows
competitive performance with a moderate parameter overhead,  
demonstrating its strong compatibility with different pretraining schemes.

\vspace{0.3em}
\noindent\textbf{Under MAE pretraining}, our method achieves competitive performance, notably outperforming prior parameter-efficient baselines such as VPT~\cite{jia2022visual}, GPT~\cite{yoo2023improving}, and VFPT in the \textit{Specialized} category (+0.26\% over VFPT), which often includes domain-specific tasks (e.g., medical, satellite images). Despite MAE’s inherent information loss due to reconstruction-based learning, our approach is able to extract and refine meaningful representations through instance-aware prompt sampling.

\vspace{0.3em}
\noindent\textbf{Under MoCo v3 pretraining}, our method consistently surpasses all baselines, achieving top results in all three VTAB-1k categories. In particular, we observe notable gains in the \textit{Structured} group (+1.31\% over VFPT), which tends to exhibit distributional shifts far from the pretraining domain. These results reveal that the instance-aware latent space introduced by our prompt encoder provides enhanced resilience to domain gaps, even under contrastive pretraining that prioritizes instance-level discrimination.

\vspace{0.3em}
\noindent\textbf{Prompt robustness to pretraining variance.} Prior studies showed that prompt-based tuning may suffer from unstable performance under self-supervised pretraining~\cite{zang2022unified, zhou2022conditional}. In contrast, our method yields \textit{consistent gains across all categories and paradigms}, suggesting that subspace-guided prompt decomposition improves the semantic reliability of learned tokens, regardless of upstream objective.

\begin{table}[!ht]
\caption{Ablation study on instance-aware prompt and PCA-based decomposition using ViT-Base/16 on VTAB-1k.}
\label{table:ablation_main}
\centering
\resizebox{0.99\linewidth}{!}{
\begin{tabular}{l|ccc|c}
\Xhline{4\arrayrulewidth}
\rowcolor{gray!20}
\textbf{Method} & \textbf{Natural} & \textbf{Specialized} & \textbf{Structured} & \textbf{VTAB-1k Mean} \\
\hline
w/o PCA & 80.72 & 83.14 & 59.18 & 74.35 \\
w/o Instance Prompt & 81.15 & 83.90 & 60.05 & 75.03 \\
w/o PCA \& Inst. & 79.68 & 81.94 & 57.92 & 73.18 \\
w/ Random Prompt & 80.10 & 82.53 & 58.70 & 73.78 \\
\rowcolor{gray!20}
\textbf{Full Model (Ours)} & \textbf{82.62} & \textbf{85.22} & \textbf{61.25} & \textbf{76.36} \\
\Xhline{4\arrayrulewidth}
\end{tabular}
}
\vspace{-0.5em}
\end{table}

\begin{figure}[t]
    \centering
    \includegraphics[width=0.48\textwidth]{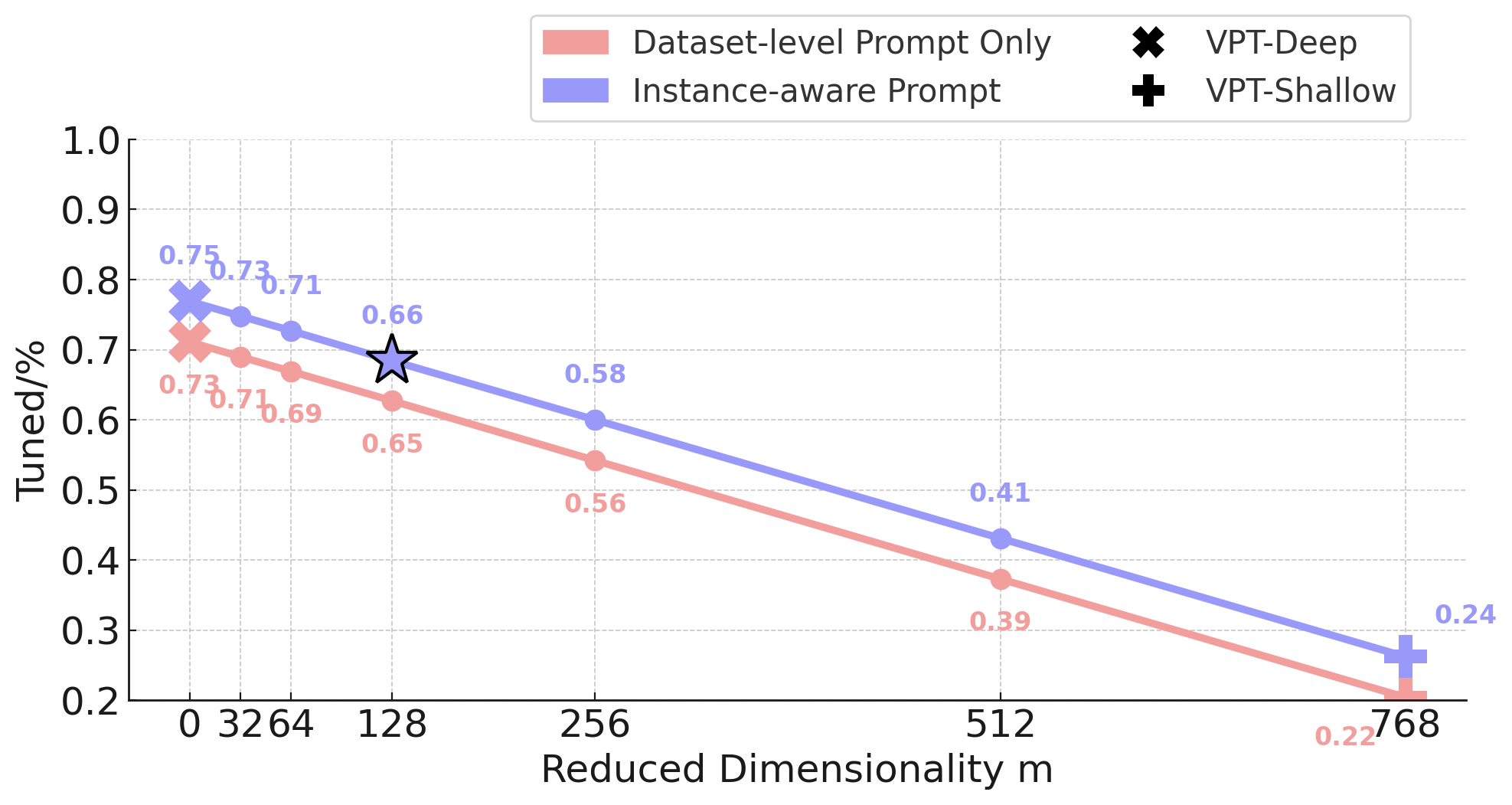}
    \caption{Parameter efficiency vs. dimensionality. Trainable/total parameter ratio across dimensionalities ($m$).}
    \label{fig:param_ratio_curve}
    \vspace{-0.5em}
\end{figure}

\subsection{Ablation Study}
\label{sec:ablation}

To understand the contribution of each core component in our framework, we conduct a comprehensive ablation study on the VTAB-1k benchmark using ViT-Base/16 pretrained on ImageNet-21k. As shown in Table~\ref{table:ablation_main}, we systematically remove or replace key modules to evaluate their individual and combined effects.

\vspace{0.3em}
\noindent\textbf{Effectiveness of instance-aware prompts.} When removing the instance-aware prompting while retaining PCA compression, the average VTAB-1k accuracy drops from 76.36\% to 75.03\%. This 1.33\% degradation confirms that static dataset-level prompts are suboptimal for handling input variability, and highlights the benefit of injecting personalized prompt information on a per-instance basis.

\vspace{0.3em}
\noindent\textbf{Trade-off between efficiency and accuracy.} 
To further explore the efficiency of our prompt strategy, we visualize the ratio of trainable parameters to the total model size across varying prompt dimensionalities in Fig.~\ref{fig:param_ratio_curve}. The instance-aware variant (blue) maintains a consistently lower parameter ratio compared to the dataset-level prompt baseline (pink), particularly when the dimensionality is reduced. This reflects the parameter advantage of our design, which offloads representation burden from prompt tokens to the lightweight instance generator. Notably, despite tuning fewer parameters, our method achieves superior or comparable accuracy at nearly every dimension (Fig.~\ref{fig:intro_curve}). These results suggest that semantic compression not only improves performance but also yields better parameter efficiency, offering a favorable trade-off in resource-limited scenarios.

\vspace{0.3em}
\noindent\textbf{Benefit of semantic compression via PCA.} As shown in Fig~\ref{fig:intro_curve}, eliminating the PCA compression while keeping instance-aware prompts results in a 2.01\% drop in average accuracy (from 76.36\% to 74.35\%). This gap demonstrates the importance of semantic bottlenecking: PCA selects task-relevant latent directions, filters noisy or redundant features, and facilitates stable prompt transfer across transformer layers.

\vspace{0.3em}
\noindent\textbf{Complementarity between components.} The most substantial performance drop (3.18\%) occurs when both components are disabled (i.e., dataset-level random prompts without PCA), resulting in 73.18\% accuracy. This indicates that PCA and instance prompts not only contribute individually but also reinforce each other: semantic compression guides prompt generation towards informative subspaces, while instance conditioning ensures adaptiveness within that subspace.


\subsection{Interpretability and Feature Representation}
\label{sec:interpretability}

To gain deeper insights into how our instance-aware and PCA-refined prompts influence the learned representations, we conduct qualitative and quantitative interpretability analyses, including Grad-CAM visualizations, t-SNE embeddings, and token similarity heatmaps. These tools help uncover the semantic structure, instance adaptivity, and representational compactness of the proposed method.

\begin{figure}[t]
    \centering
    \includegraphics[width=0.3\textwidth]{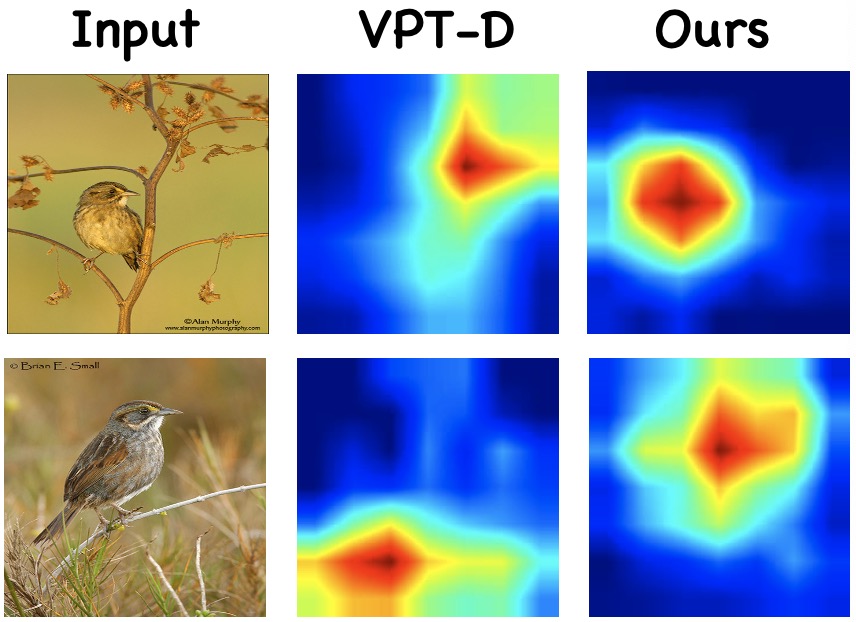}
    \caption{Grad-CAM heatmaps on CUB-200 samples. Compared to VPT-Deep, ViaPT generates more focused and semantically aligned heatmaps, demonstrating better spatial consistency with object regions.}
    \label{fig:gradcam}
    \vspace{-0.5em}
\end{figure}

\vspace{0.5em}
\noindent\textbf{Heatmap Visualization via Grad-CAM.}
To explore how prompt design influences attention allocation, we visualize class-specific activation heatmaps using Grad-CAM~\cite{selvaraju2017grad} over the [CLS] token in the final transformer layer. As shown in Fig.~\ref{fig:gradcam}, VPT-Deep often exhibits diffused or off-target activation, failing to focus on semantically relevant regions. In contrast, our instance-aware prompt generation produces sharper and more centralized heatmaps, consistently aligning with the object of interest (e.g., the bird body or head), indicating improved feature localization and interpretability.

\begin{figure}[t]
    \centering
    \includegraphics[width=0.45\textwidth]{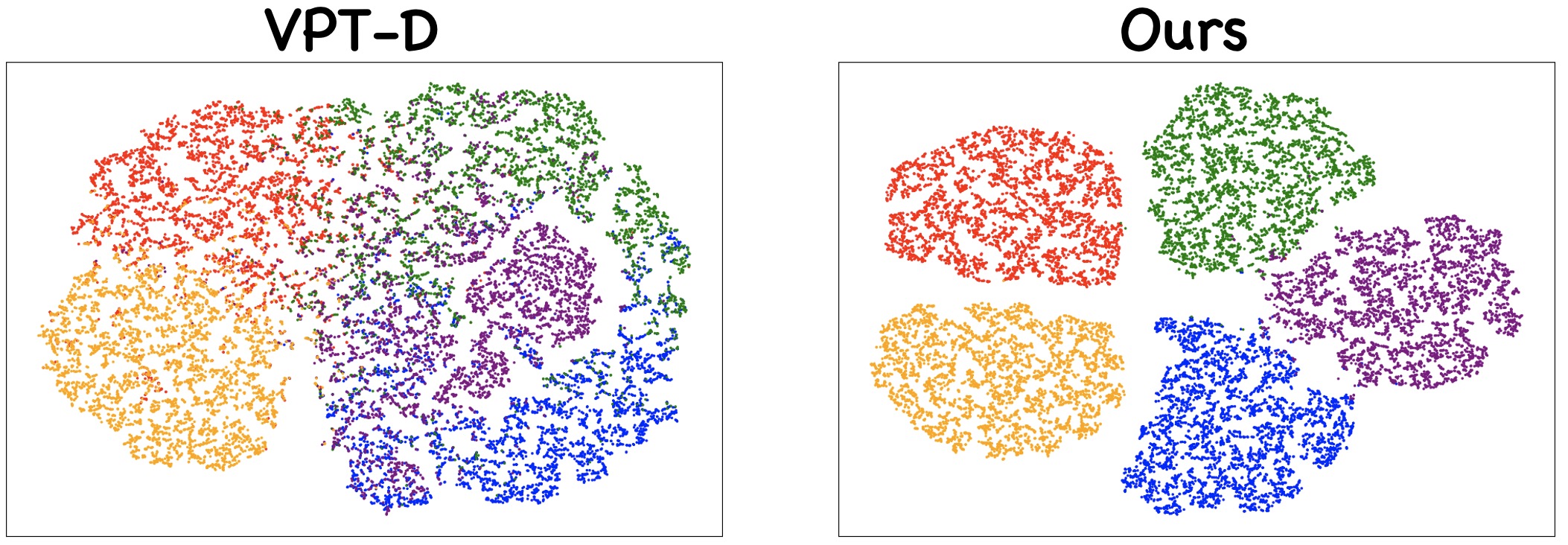}
    \caption{t-SNE visualization on CUB-200. The proposed ViaPT yields compact and well-separated clusters across five categories, demonstrating enhanced semantic discriminability compared to VPT-Deep.}
    \label{fig:tsne}
    \vspace{-0.5em}
\end{figure}

\vspace{0.5em}
\noindent\textbf{t-SNE Embedding Analysis.}  
To assess the effectiveness of our instance-aware prompt learning from a representational perspective, we visualize the output class tokens $\mathbf{x}_N$ using t-SNE~\cite{van2008visualizing}. Fig.~\ref{fig:tsne} presents a comparative visualization of the embeddings produced by VPT-Deep and our ViaPT on the CUB-200 dataset with five randomly sampled categories. The embeddings derived from our method exhibit markedly improved cluster compactness and inter-class separability, forming clearly delineated and semantically structured regions in the latent space. This suggests that instance-specific prompt modeling contributes to more consistent intra-class alignment and stronger discriminative capability. In contrast, VPT-Deep embeddings show significant overlap between classes, indicating weaker category specificity and feature confusion. These results corroborate the hypothesis that our proposed framework facilitates superior semantic organization in the feature space.

\begin{figure}[t]
    \centering
    \includegraphics[width=0.3\textwidth]{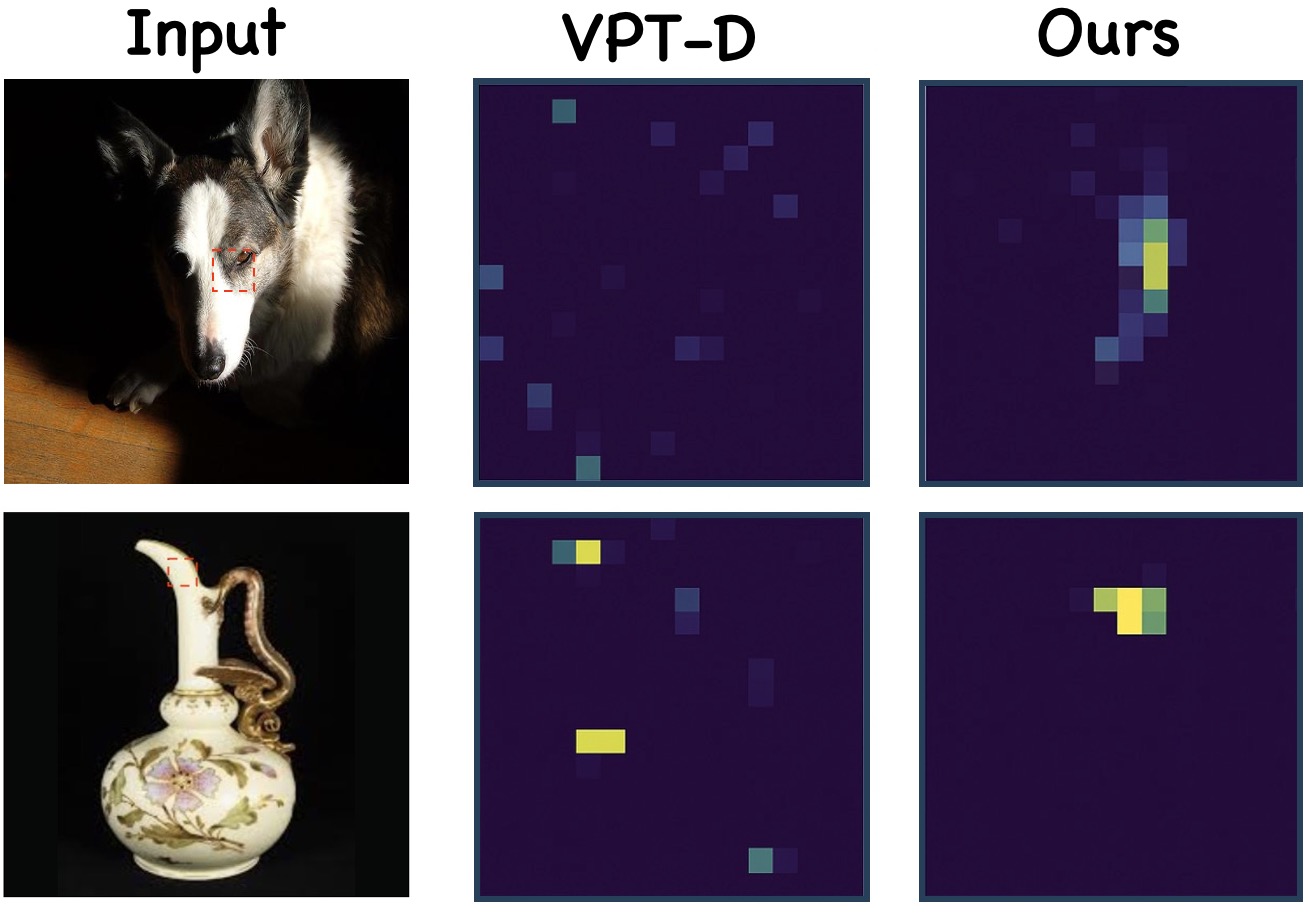}
    \caption{\textbf{Prompt token similarity heatmaps.} Compared with VPT-Deep, our instance-aware prompts yield more focused and semantically meaningful attention maps. }
    \label{fig:similarity}
    \vspace{-0.5em}
\end{figure}

\vspace{0.5em}
\noindent\textbf{Prompt Token Similarity Visualization.}
To further understand the effectiveness of our instance-aware prompt design, we visualize the prompt token similarity maps for selected images. Specifically, we compute cosine similarity between prompt tokens and spatial locations at the first transformer layer. As illustrated in Fig.~\ref{fig:similarity}, VPT-Deep produces highly sparse and noisy activations, with attention scattered across background regions. In contrast, our method exhibits more concentrated and semantically meaningful attention patterns—focusing on task-relevant parts such as the dog’s face and the vase’s spout. This suggests that our instance-aware prompt generation effectively guides attention to critical visual cues, leading to improved prompt utility and reduced redundancy.

\section{Conclusion}

We reveal that instance-aware prompts can significantly boost prompt tuning for vision transformers. This is achieved by the proposed ViaPT method. We show that the key to ViaPT includes generating instance-aware prompts based on each individual input and leveraging PCA to retain important prompting information when fusing dataset-level and instance-aware prompts. Based our proposed conceptual understanding, the relationship between ViaPT and typical VPT schemes is interpreted, indicating the rationale of our method from another perspective. The promising results on diverse and challenging experiments demonstrate the competitiveness of ViaPT, potentially benefiting the research communities of deep transfer learning and  foundation models. 





\begin{acks}
This work was supported in part by the U.S. NIH grant R01GM134020.  
This manuscript was authored by Oak Ridge National Laboratory (ORNL), operated by UT-Battelle, LLC under Contract No. DE-AC05-00OR22725 with the U.S. Department of Energy.  
Any subjective views or opinions expressed in this paper do not necessarily represent those of the U.S. Department of Energy or the United States Government.
\end{acks}

\bibliographystyle{ACM-Reference-Format}
\bibliography{sample-base}


\begin{thebibliography}{86}


\ifx \showCODEN    \undefined \def \showCODEN     #1{\unskip}     \fi
\ifx \showISBNx    \undefined \def \showISBNx     #1{\unskip}     \fi
\ifx \showISBNxiii \undefined \def \showISBNxiii  #1{\unskip}     \fi
\ifx \showISSN     \undefined \def \showISSN      #1{\unskip}     \fi
\ifx \showLCCN     \undefined \def \showLCCN      #1{\unskip}     \fi
\ifx \shownote     \undefined \def \shownote      #1{#1}          \fi
\ifx \showarticletitle \undefined \def \showarticletitle #1{#1}   \fi
\ifx \showURL      \undefined \def \showURL       {\relax}        \fi
\providecommand\bibfield[2]{#2}
\providecommand\bibinfo[2]{#2}
\providecommand\natexlab[1]{#1}
\providecommand\showeprint[2][]{arXiv:#2}

\bibitem[Achtibat et~al\mbox{.}(2023)]%
        {achtibat2023attribution}
\bibfield{author}{\bibinfo{person}{Reduan Achtibat}, \bibinfo{person}{Maximilian Dreyer}, \bibinfo{person}{Ilona Eisenbraun}, \bibinfo{person}{Sebastian Bosse}, \bibinfo{person}{Thomas Wiegand}, \bibinfo{person}{Wojciech Samek}, {and} \bibinfo{person}{Sebastian Lapuschkin}.} \bibinfo{year}{2023}\natexlab{}.
\newblock \showarticletitle{From attribution maps to human-understandable explanations through concept relevance propagation}.
\newblock \bibinfo{journal}{\emph{Nature Machine Intelligence}} \bibinfo{volume}{5}, \bibinfo{number}{9} (\bibinfo{year}{2023}), \bibinfo{pages}{1006--1019}.
\newblock


\bibitem[Ba et~al\mbox{.}(2016)]%
        {ba2016layer}
\bibfield{author}{\bibinfo{person}{Jimmy~Lei Ba}, \bibinfo{person}{Jamie~Ryan Kiros}, {and} \bibinfo{person}{Geoffrey~E Hinton}.} \bibinfo{year}{2016}\natexlab{}.
\newblock \showarticletitle{Layer normalization}.
\newblock \bibinfo{journal}{\emph{arXiv preprint arXiv:1607.06450}} (\bibinfo{year}{2016}).
\newblock


\bibitem[Bahng et~al\mbox{.}(2022a)]%
        {bahng2022exploring}
\bibfield{author}{\bibinfo{person}{Hyojin Bahng}, \bibinfo{person}{Ali Jahanian}, \bibinfo{person}{Swami Sankaranarayanan}, {and} \bibinfo{person}{Phillip Isola}.} \bibinfo{year}{2022}\natexlab{a}.
\newblock \bibinfo{title}{Exploring visual prompts for adapting large-scale models}.
\newblock


\bibitem[Bahng et~al\mbox{.}(2022b)]%
        {bahng2022visual}
\bibfield{author}{\bibinfo{person}{Hyojin Bahng}, \bibinfo{person}{Ali Jahanian}, \bibinfo{person}{Swami Sankaranarayanan}, {and} \bibinfo{person}{Phillip Isola}.} \bibinfo{year}{2022}\natexlab{b}.
\newblock \showarticletitle{Visual prompting: Modifying pixel space to adapt pre-trained models}.
\newblock \bibinfo{journal}{\emph{arXiv preprint arXiv:2203.17274}} (\bibinfo{year}{2022}).
\newblock


\bibitem[Bengio et~al\mbox{.}(2015)]%
        {bengio2015conditional}
\bibfield{author}{\bibinfo{person}{Emmanuel Bengio}, \bibinfo{person}{Pierre-Luc Bacon}, \bibinfo{person}{Joelle Pineau}, {and} \bibinfo{person}{Doina Precup}.} \bibinfo{year}{2015}\natexlab{}.
\newblock \showarticletitle{Conditional computation in neural networks for faster models}.
\newblock \bibinfo{journal}{\emph{arXiv preprint arXiv:1511.06297}} (\bibinfo{year}{2015}).
\newblock


\bibitem[Bossard et~al\mbox{.}(2014)]%
        {bossard2014food}
\bibfield{author}{\bibinfo{person}{Lukas Bossard}, \bibinfo{person}{Matthieu Guillaumin}, {and} \bibinfo{person}{Luc Van~Gool}.} \bibinfo{year}{2014}\natexlab{}.
\newblock \showarticletitle{Food-101--mining discriminative components with random forests}. In \bibinfo{booktitle}{\emph{ECCV}}.
\newblock


\bibitem[Cai et~al\mbox{.}(2020)]%
        {cai2020tinytl}
\bibfield{author}{\bibinfo{person}{Han Cai}, \bibinfo{person}{Chuang Gan}, \bibinfo{person}{Ligeng Zhu}, {and} \bibinfo{person}{Song Han}.} \bibinfo{year}{2020}\natexlab{}.
\newblock \bibinfo{title}{Tinytl: Reduce memory, not parameters for efficient on-device learning}.
\newblock


\bibitem[Chen et~al\mbox{.}(2022)]%
        {chen2022adaptformer}
\bibfield{author}{\bibinfo{person}{Shoufa Chen}, \bibinfo{person}{Chuang Ge}, \bibinfo{person}{Zhiqiang Tong}, \bibinfo{person}{Jianzhuang Wang}, \bibinfo{person}{Yang Song}, \bibinfo{person}{Jianmin Wang}, {and} \bibinfo{person}{Ping Luo}.} \bibinfo{year}{2022}\natexlab{}.
\newblock \showarticletitle{Adaptformer: Adapting vision transformers for scalable visual recognition}.
\newblock \bibinfo{journal}{\emph{arXiv preprint arXiv:2205.13535}} (\bibinfo{year}{2022}).
\newblock


\bibitem[Chen et~al\mbox{.}(2020)]%
        {chen2020improved}
\bibfield{author}{\bibinfo{person}{Xinlei Chen}, \bibinfo{person}{Haoqi Fan}, \bibinfo{person}{Ross Girshick}, {and} \bibinfo{person}{Kaiming He}.} \bibinfo{year}{2020}\natexlab{}.
\newblock \bibinfo{title}{Improved baselines with momentum contrastive learning}.
\newblock


\bibitem[Chen et~al\mbox{.}(2021)]%
        {chen2021empirical}
\bibfield{author}{\bibinfo{person}{Xinlei Chen}, \bibinfo{person}{Saining Xie}, {and} \bibinfo{person}{Kaiming He}.} \bibinfo{year}{2021}\natexlab{}.
\newblock \bibinfo{title}{An empirical study of training self-supervised vision transformers}.
\newblock


\bibitem[Cheng et~al\mbox{.}(2023)]%
        {cheng2023e2vpt}
\bibfield{author}{\bibinfo{person}{Han Cheng}, \bibinfo{person}{Wang Qifan}, \bibinfo{person}{Cui Yiming}, \bibinfo{person}{Cao Zhiwen}, \bibinfo{person}{Wang Wenguan}, \bibinfo{person}{Qi Siyuan}, {and} \bibinfo{person}{Liu Dongfang}.} \bibinfo{year}{2023}\natexlab{}.
\newblock \showarticletitle{E2VPT: An Effective and Efficient Approach for Visual Prompt Tuning}. In \bibinfo{booktitle}{\emph{International Conference on Computer Vision (ICCV)}}.
\newblock


\bibitem[Chowdhury et~al\mbox{.}(2025)]%
        {Chowdhury_2025_CVPR}
\bibfield{author}{\bibinfo{person}{Arpita Chowdhury}, \bibinfo{person}{Dipanjyoti Paul}, \bibinfo{person}{Zheda Mai}, \bibinfo{person}{Jianyang Gu}, \bibinfo{person}{Ziheng Zhang}, \bibinfo{person}{Kazi~Sajeed Mehrab}, \bibinfo{person}{Elizabeth~G. Campolongo}, \bibinfo{person}{Daniel Rubenstein}, \bibinfo{person}{Charles~V. Stewart}, \bibinfo{person}{Anuj Karpatne}, \bibinfo{person}{Tanya Berger-Wolf}, \bibinfo{person}{Yu Su}, {and} \bibinfo{person}{Wei-Lun Chao}.} \bibinfo{year}{2025}\natexlab{}.
\newblock \showarticletitle{Prompt-CAM: Making Vision Transformers Interpretable for Fine-Grained Analysis}. In \bibinfo{booktitle}{\emph{Proceedings of the Computer Vision and Pattern Recognition Conference (CVPR)}}. \bibinfo{pages}{4375--4385}.
\newblock


\bibitem[Cimpoi et~al\mbox{.}(2014)]%
        {cimpoi2014describing}
\bibfield{author}{\bibinfo{person}{Mircea Cimpoi}, \bibinfo{person}{Subhransu Maji}, \bibinfo{person}{Iasonas Kokkinos}, \bibinfo{person}{Sammy Mohamed}, {and} \bibinfo{person}{Andrea Vedaldi}.} \bibinfo{year}{2014}\natexlab{}.
\newblock \showarticletitle{Describing textures in the wild}. In \bibinfo{booktitle}{\emph{CVPR}}.
\newblock


\bibitem[Das et~al\mbox{.}(2023)]%
        {das2023learning}
\bibfield{author}{\bibinfo{person}{Rahul Das}, \bibinfo{person}{Yahel Dukler}, \bibinfo{person}{Avinash Ravichandran}, {and} \bibinfo{person}{Ajay Swaminathan}.} \bibinfo{year}{2023}\natexlab{}.
\newblock \showarticletitle{Learning Expressive Prompting With Residuals for Vision Transformers}. In \bibinfo{booktitle}{\emph{CVPR}}.
\newblock


\bibitem[Deng et~al\mbox{.}(2009)]%
        {deng2009imagenet}
\bibfield{author}{\bibinfo{person}{Jia Deng}, \bibinfo{person}{Wei Dong}, \bibinfo{person}{Richard Socher}, \bibinfo{person}{Li-Jia Li}, \bibinfo{person}{Kai Li}, {and} \bibinfo{person}{Li Fei-Fei}.} \bibinfo{year}{2009}\natexlab{}.
\newblock \showarticletitle{Imagenet: A large-scale hierarchical image database}. In \bibinfo{booktitle}{\emph{CVPR}}.
\newblock


\bibitem[Dong et~al\mbox{.}(2023)]%
        {dong2023efficient}
\bibfield{author}{\bibinfo{person}{Shaohua Dong}, \bibinfo{person}{Yunhe Feng}, \bibinfo{person}{Qing Yang}, \bibinfo{person}{Yan Huang}, \bibinfo{person}{Dongfang Liu}, {and} \bibinfo{person}{Heng Fan}.} \bibinfo{year}{2023}\natexlab{}.
\newblock \bibinfo{title}{Efficient Multimodal Semantic Segmentation via Dual-Prompt Learning}.
\newblock


\bibitem[Dosovitskiy et~al\mbox{.}(2020)]%
        {dosovitskiy2020image}
\bibfield{author}{\bibinfo{person}{Alexey Dosovitskiy}, \bibinfo{person}{Lucas Beyer}, \bibinfo{person}{Alexander Kolesnikov}, \bibinfo{person}{Dirk Weissenborn}, \bibinfo{person}{Xiaohua Zhai}, \bibinfo{person}{Thomas Unterthiner}, \bibinfo{person}{Mostafa Dehghani}, \bibinfo{person}{Matthias Minderer}, \bibinfo{person}{Georg Heigold}, \bibinfo{person}{Sylvain Gelly}, {et~al\mbox{.}}} \bibinfo{year}{2020}\natexlab{}.
\newblock \showarticletitle{An image is worth 16x16 words: Transformers for image recognition at scale}.
\newblock \bibinfo{journal}{\emph{arXiv preprint arXiv:2010.11929}} (\bibinfo{year}{2020}).
\newblock


\bibitem[Gebru et~al\mbox{.}(2017)]%
        {gebru2017stanfordcars}
\bibfield{author}{\bibinfo{person}{Timnit Gebru}, \bibinfo{person}{Jonathan Krause}, \bibinfo{person}{Yilun Wang}, \bibinfo{person}{Duyun Chen}, \bibinfo{person}{Jia Deng}, \bibinfo{person}{Erez~Lieberman Aiden}, {and} \bibinfo{person}{Li Fei-Fei}.} \bibinfo{year}{2017}\natexlab{}.
\newblock \showarticletitle{Fine-grained car detection for visual census estimation}. In \bibinfo{booktitle}{\emph{AAAI Conference on Artificial Intelligence}}.
\newblock


\bibitem[Goswami et~al\mbox{.}(2024)]%
        {goswami2024copl}
\bibfield{author}{\bibinfo{person}{Koustava Goswami}, \bibinfo{person}{Srikrishna Karanam}, \bibinfo{person}{Prateksha Udhayanan}, \bibinfo{person}{KJ Joseph}, {and} \bibinfo{person}{Balaji~Vasan Srinivasan}.} \bibinfo{year}{2024}\natexlab{}.
\newblock \showarticletitle{Copl: Contextual prompt learning for vision-language understanding}. In \bibinfo{booktitle}{\emph{Proceedings of the AAAI Conference on Artificial Intelligence}}, Vol.~\bibinfo{volume}{38}. \bibinfo{pages}{18090--18098}.
\newblock


\bibitem[Han et~al\mbox{.}(2023b)]%
        {han20232vpt}
\bibfield{author}{\bibinfo{person}{Cheng Han}, \bibinfo{person}{Qifan Wang}, \bibinfo{person}{Yiming Cui}, \bibinfo{person}{Zhiwen Cao}, \bibinfo{person}{Wenguan Wang}, \bibinfo{person}{Siyuan Qi}, {and} \bibinfo{person}{Dongfang Liu}.} \bibinfo{year}{2023}\natexlab{b}.
\newblock \bibinfo{title}{E2vpt: An effective and efficient approach for visual prompt tuning}.
\newblock


\bibitem[Han et~al\mbox{.}(2023c)]%
        {han2023e2vpt}
\bibfield{author}{\bibinfo{person}{Chunyuan Han}, \bibinfo{person}{Qi Wang}, \bibinfo{person}{Yuxin Cui}, \bibinfo{person}{Zhiqian Cao}, \bibinfo{person}{Wenqing Wang}, \bibinfo{person}{Shiyu Qi}, {and} \bibinfo{person}{Ding Liu}.} \bibinfo{year}{2023}\natexlab{c}.
\newblock \showarticletitle{Eˆ 2VPT: An Effective and Efficient Approach for Visual Prompt Tuning}.
\newblock \bibinfo{journal}{\emph{arXiv preprint arXiv:2307.13770}} (\bibinfo{year}{2023}).
\newblock


\bibitem[Han et~al\mbox{.}(2024)]%
        {han2024facing}
\bibfield{author}{\bibinfo{person}{Cheng Han}, \bibinfo{person}{Qifan Wang}, \bibinfo{person}{Yiming Cui}, \bibinfo{person}{Wenguan Wang}, \bibinfo{person}{Lifu Huang}, {and} \bibinfo{person}{Dongfang Liu}.} \bibinfo{year}{2024}\natexlab{}.
\newblock \showarticletitle{Facing the Elephant in the Room: Visual Prompt Tuning or Full Finetuning?}. In \bibinfo{booktitle}{\emph{International Conference on Learning Representations (ICLR)}}.
\newblock


\bibitem[Han et~al\mbox{.}(2023a)]%
        {han2023flatten}
\bibfield{author}{\bibinfo{person}{Dongchen Han}, \bibinfo{person}{Xuran Pan}, \bibinfo{person}{Yizeng Han}, \bibinfo{person}{Shiji Song}, {and} \bibinfo{person}{Gao Huang}.} \bibinfo{year}{2023}\natexlab{a}.
\newblock \showarticletitle{Flatten transformer: Vision transformer using focused linear attention}. In \bibinfo{booktitle}{\emph{Proceedings of the IEEE/CVF international conference on computer vision}}. \bibinfo{pages}{5961--5971}.
\newblock


\bibitem[He et~al\mbox{.}(2022)]%
        {he2022masked}
\bibfield{author}{\bibinfo{person}{Kaiming He}, \bibinfo{person}{Xinlei Chen}, \bibinfo{person}{Saining Xie}, \bibinfo{person}{Yanghao Li}, \bibinfo{person}{Piotr Doll{\'a}r}, {and} \bibinfo{person}{Ross Girshick}.} \bibinfo{year}{2022}\natexlab{}.
\newblock \bibinfo{title}{Masked autoencoders are scalable vision learners}.
\newblock


\bibitem[He et~al\mbox{.}(2016)]%
        {he2016deep}
\bibfield{author}{\bibinfo{person}{Kaiming He}, \bibinfo{person}{Xiangyu Zhang}, \bibinfo{person}{Shaoqing Ren}, {and} \bibinfo{person}{Jian Sun}.} \bibinfo{year}{2016}\natexlab{}.
\newblock \showarticletitle{Deep residual learning for image recognition}. In \bibinfo{booktitle}{\emph{CVPR}}.
\newblock


\bibitem[Helber et~al\mbox{.}(2019)]%
        {helber2019eurosat}
\bibfield{author}{\bibinfo{person}{Patrick Helber}, \bibinfo{person}{Benjamin Bischke}, \bibinfo{person}{Andreas Dengel}, {and} \bibinfo{person}{Damian Borth}.} \bibinfo{year}{2019}\natexlab{}.
\newblock \showarticletitle{Eurosat: A novel dataset and deep learning benchmark for land use and land cover classification}.
\newblock \bibinfo{journal}{\emph{IEEE Journal of Selected Topics in Applied Earth Observations and Remote Sensing}} \bibinfo{volume}{12}, \bibinfo{number}{7} (\bibinfo{year}{2019}), \bibinfo{pages}{2217--2226}.
\newblock


\bibitem[Higgins et~al\mbox{.}(2017)]%
        {higgins2017beta}
\bibfield{author}{\bibinfo{person}{Irina Higgins}, \bibinfo{person}{Loic Matthey}, \bibinfo{person}{Arka Pal}, \bibinfo{person}{Christopher Burgess}, \bibinfo{person}{Xavier Glorot}, \bibinfo{person}{Matthew Botvinick}, \bibinfo{person}{Shakir Mohamed}, {and} \bibinfo{person}{Alexander Lerchner}.} \bibinfo{year}{2017}\natexlab{}.
\newblock \showarticletitle{beta-vae: Learning basic visual concepts with a constrained variational framework}. In \bibinfo{booktitle}{\emph{International conference on learning representations}}.
\newblock


\bibitem[Houlsby et~al\mbox{.}(2019)]%
        {houlsby2019parameter}
\bibfield{author}{\bibinfo{person}{Neil Houlsby} {et~al\mbox{.}}} \bibinfo{year}{2019}\natexlab{}.
\newblock \showarticletitle{Parameter-efficient transfer learning for NLP}. In \bibinfo{booktitle}{\emph{ICML}}.
\newblock


\bibitem[Hu et~al\mbox{.}(2022)]%
        {hu2022lora}
\bibfield{author}{\bibinfo{person}{Edward~J Hu}, \bibinfo{person}{Yelong Shen}, \bibinfo{person}{Phillip Wallis}, \bibinfo{person}{Zeyuan Allen-Zhu}, \bibinfo{person}{Yuanzhi Li}, \bibinfo{person}{Shean Wang}, \bibinfo{person}{Lu Wang}, {and} \bibinfo{person}{Weizhu Chen}.} \bibinfo{year}{2022}\natexlab{}.
\newblock \showarticletitle{LoRA: Low-Rank Adaptation of Large Language Models}. In \bibinfo{booktitle}{\emph{ICLR}}.
\newblock


\bibitem[Huang et~al\mbox{.}(2017)]%
        {huang2017densely}
\bibfield{author}{\bibinfo{person}{Gao Huang}, \bibinfo{person}{Zhuang Liu}, \bibinfo{person}{Laurens Van Der~Maaten}, {and} \bibinfo{person}{Kilian~Q Weinberger}.} \bibinfo{year}{2017}\natexlab{}.
\newblock \showarticletitle{Densely connected convolutional networks}. In \bibinfo{booktitle}{\emph{Proceedings of the IEEE Conference on Computer Vision and Pattern Recognition (CVPR)}}. \bibinfo{pages}{4700--4708}.
\newblock


\bibitem[Huang et~al\mbox{.}(2023b)]%
        {huang2023hta}
\bibfield{author}{\bibinfo{person}{Mingzhen Huang}, \bibinfo{person}{Jingru Zhang}, \bibinfo{person}{Xiaodan Liang}, {and} \bibinfo{person}{Hao Wang}.} \bibinfo{year}{2023}\natexlab{b}.
\newblock \showarticletitle{DAM-VP: Adaptive Meta-Learning for Visual Prompt Tuning in Domain-Adaptive Vision Transformers}. In \bibinfo{booktitle}{\emph{Proceedings of the IEEE Conference on Computer Vision and Pattern Recognition}}.
\newblock


\bibitem[Huang et~al\mbox{.}(2023a)]%
        {huang2023diversity}
\bibfield{author}{\bibinfo{person}{Qingyao Huang}, \bibinfo{person}{Xingchen Dong}, \bibinfo{person}{Dantong Chen}, \bibinfo{person}{Weiwei Zhang}, \bibinfo{person}{Fuzhen Wang}, \bibinfo{person}{Gang Hua}, {and} \bibinfo{person}{Nenghai Yu}.} \bibinfo{year}{2023}\natexlab{a}.
\newblock \showarticletitle{Diversity-Aware Meta Visual Prompting}. In \bibinfo{booktitle}{\emph{CVPR}}.
\newblock


\bibitem[Iofinova et~al\mbox{.}(2022)]%
        {iofinova2022well}
\bibfield{author}{\bibinfo{person}{Eugenia Iofinova}, \bibinfo{person}{Alexandra Peste}, \bibinfo{person}{Mark Kurtz}, {and} \bibinfo{person}{Dan Alistarh}.} \bibinfo{year}{2022}\natexlab{}.
\newblock \showarticletitle{How well do sparse imagenet models transfer?}. In \bibinfo{booktitle}{\emph{Proceedings of the IEEE/CVF Conference on Computer Vision and Pattern Recognition}}. \bibinfo{pages}{12266--12276}.
\newblock


\bibitem[Ji et~al\mbox{.}(2025)]%
        {ji2025cibrcrossmodalinformationbottleneck}
\bibfield{author}{\bibinfo{person}{Yingrui Ji}, \bibinfo{person}{Xi Xiao}, \bibinfo{person}{Gaofei Chen}, \bibinfo{person}{Hao Xu}, \bibinfo{person}{Chenrui Ma}, \bibinfo{person}{Lijing Zhu}, \bibinfo{person}{Aokun Liang}, {and} \bibinfo{person}{Jiansheng Chen}.} \bibinfo{year}{2025}\natexlab{}.
\newblock \bibinfo{title}{CIBR: Cross-modal Information Bottleneck Regularization for Robust CLIP Generalization}.
\newblock
\showeprint[arxiv]{2503.24182}~[cs.CV]
\urldef\tempurl%
\url{https://arxiv.org/abs/2503.24182}
\showURL{%
\tempurl}


\bibitem[Jia et~al\mbox{.}(2022)]%
        {jia2022visual}
\bibfield{author}{\bibinfo{person}{Menglin Jia}, \bibinfo{person}{Luming Tang}, \bibinfo{person}{Bor-Chun Chen}, \bibinfo{person}{Claire Cardie}, \bibinfo{person}{Serge Belongie}, \bibinfo{person}{Bharath Hariharan}, {and} \bibinfo{person}{Ser-Nam Lim}.} \bibinfo{year}{2022}\natexlab{}.
\newblock \showarticletitle{Visual prompt tuning}. In \bibinfo{booktitle}{\emph{ECCV}}.
\newblock


\bibitem[Jin et~al\mbox{.}({[n.\,d.]})]%
        {jinlor}
\bibfield{author}{\bibinfo{person}{Can Jin}, \bibinfo{person}{Ying Li}, \bibinfo{person}{Mingyu Zhao}, \bibinfo{person}{Shiyu Zhao}, \bibinfo{person}{Zhenting Wang}, \bibinfo{person}{Xiaoxiao He}, \bibinfo{person}{Ligong Han}, \bibinfo{person}{Tong Che}, {and} \bibinfo{person}{Dimitris~N Metaxas}.} \bibinfo{year}{[n.\,d.]}\natexlab{}.
\newblock \showarticletitle{LoR-VP: Low-Rank Visual Prompting for Efficient Vision Model Adaptation}. In \bibinfo{booktitle}{\emph{The Thirteenth International Conference on Learning Representations}}.
\newblock


\bibitem[Khosla et~al\mbox{.}(2011)]%
        {khosla2011stanforddogs}
\bibfield{author}{\bibinfo{person}{Aditya Khosla}, \bibinfo{person}{Nityananda Jayadevaprakash}, \bibinfo{person}{Bangpeng Yao}, {and} \bibinfo{person}{Li Fei-Fei}.} \bibinfo{year}{2011}\natexlab{}.
\newblock \showarticletitle{Novel dataset for fine-grained image categorization}. In \bibinfo{booktitle}{\emph{Proceedings of the IEEE Conference on Computer Vision and Pattern Recognition}}. \bibinfo{pages}{109--116}.
\newblock


\bibitem[Kingma and Welling(2013)]%
        {kingma2013auto}
\bibfield{author}{\bibinfo{person}{Diederik~P Kingma} {and} \bibinfo{person}{Max Welling}.} \bibinfo{year}{2013}\natexlab{}.
\newblock \showarticletitle{Auto-Encoding Variational Bayes}.
\newblock \bibinfo{journal}{\emph{arXiv preprint arXiv:1312.6114}} (\bibinfo{year}{2013}).
\newblock


\bibitem[Krizhevsky and Hinton(2009)]%
        {krizhevsky2009cifar}
\bibfield{author}{\bibinfo{person}{Alex Krizhevsky} {and} \bibinfo{person}{Geoffrey Hinton}.} \bibinfo{year}{2009}\natexlab{}.
\newblock \showarticletitle{Learning multiple layers of features from tiny images}. In \bibinfo{booktitle}{\emph{Technical report, University of Toronto}}.
\newblock


\bibitem[Lester et~al\mbox{.}(2021)]%
        {lester2021power}
\bibfield{author}{\bibinfo{person}{Brian Lester}, \bibinfo{person}{Rami Al-Rfou}, {and} \bibinfo{person}{Noah Constant}.} \bibinfo{year}{2021}\natexlab{}.
\newblock \showarticletitle{The power of scale for parameter-efficient prompt tuning}.
\newblock \bibinfo{journal}{\emph{arXiv preprint arXiv:2104.08691}} (\bibinfo{year}{2021}).
\newblock


\bibitem[Li et~al\mbox{.}(2025)]%
        {li2025magicid}
\bibfield{author}{\bibinfo{person}{Hengjia Li}, \bibinfo{person}{Lifan Jiang}, \bibinfo{person}{Xi Xiao}, \bibinfo{person}{Tianyang Wang}, \bibinfo{person}{Hongwei Yi}, \bibinfo{person}{Boxi Wu}, {and} \bibinfo{person}{Deng Cai}.} \bibinfo{year}{2025}\natexlab{}.
\newblock \showarticletitle{MagicID: Hybrid Preference Optimization for ID-Consistent and Dynamic-Preserved Video Customization}.
\newblock \bibinfo{journal}{\emph{arXiv preprint arXiv:2503.12689}} (\bibinfo{year}{2025}).
\newblock


\bibitem[Li et~al\mbox{.}(2024)]%
        {li2024gca}
\bibfield{author}{\bibinfo{person}{Hengjia Li}, \bibinfo{person}{Yang Liu}, \bibinfo{person}{Yibo Zhao}, \bibinfo{person}{Haoran Cheng}, \bibinfo{person}{Yang Yang}, \bibinfo{person}{Linxuan Xia}, \bibinfo{person}{Zekai Luo}, \bibinfo{person}{Qibo Qiu}, \bibinfo{person}{Boxi Wu}, \bibinfo{person}{Tu Zheng}, {et~al\mbox{.}}} \bibinfo{year}{2024}\natexlab{}.
\newblock \showarticletitle{GCA-3D: Towards Generalized and Consistent Domain Adaptation of 3D Generators}.
\newblock \bibinfo{journal}{\emph{arXiv preprint arXiv:2412.15491}} (\bibinfo{year}{2024}).
\newblock


\bibitem[Li et~al\mbox{.}(2021)]%
        {li2021learning}
\bibfield{author}{\bibinfo{person}{Wanhua Li}, \bibinfo{person}{Xiaoke Huang}, \bibinfo{person}{Jiwen Lu}, \bibinfo{person}{Jianjiang Feng}, {and} \bibinfo{person}{Jie Zhou}.} \bibinfo{year}{2021}\natexlab{}.
\newblock \showarticletitle{Learning probabilistic ordinal embeddings for uncertainty-aware regression}. In \bibinfo{booktitle}{\emph{Proceedings of the IEEE/CVF conference on computer vision and pattern recognition}}. \bibinfo{pages}{13896--13905}.
\newblock


\bibitem[Liu and Deng(2018)]%
        {liu2018dynamic}
\bibfield{author}{\bibinfo{person}{Lanlan Liu} {and} \bibinfo{person}{Jia Deng}.} \bibinfo{year}{2018}\natexlab{}.
\newblock \showarticletitle{Dynamic deep neural networks: Optimizing accuracy-efficiency trade-offs by selective execution}. In \bibinfo{booktitle}{\emph{Proceedings of the AAAI conference on artificial intelligence}}, Vol.~\bibinfo{volume}{32}.
\newblock


\bibitem[Liu et~al\mbox{.}(2021)]%
        {liu2021swin}
\bibfield{author}{\bibinfo{person}{Ze Liu}, \bibinfo{person}{Yutong Lin}, \bibinfo{person}{Yutong Cao}, \bibinfo{person}{Han Hu}, \bibinfo{person}{Yixuan Wei}, \bibinfo{person}{Zheng Zhang}, \bibinfo{person}{Stephen Lin}, {and} \bibinfo{person}{Baining Guo}.} \bibinfo{year}{2021}\natexlab{}.
\newblock \showarticletitle{Swin transformer: Hierarchical vision transformer using shifted windows}. In \bibinfo{booktitle}{\emph{ICCV}}.
\newblock


\bibitem[Liu et~al\mbox{.}(2024)]%
        {liu2024insvp}
\bibfield{author}{\bibinfo{person}{Zichen Liu}, \bibinfo{person}{Yuxin Peng}, {and} \bibinfo{person}{Jiahuan Zhou}.} \bibinfo{year}{2024}\natexlab{}.
\newblock \showarticletitle{InsVP: Efficient Instance Visual Prompting from Image Itself}. In \bibinfo{booktitle}{\emph{Proceedings of the 32nd ACM International Conference on Multimedia}}. \bibinfo{pages}{6443--6452}.
\newblock


\bibitem[Mai et~al\mbox{.}(2025)]%
        {Mai_2025_CVPR}
\bibfield{author}{\bibinfo{person}{Zheda Mai}, \bibinfo{person}{Ping Zhang}, \bibinfo{person}{Cheng-Hao Tu}, \bibinfo{person}{Hong-You Chen}, \bibinfo{person}{Quang-Huy Nguyen}, \bibinfo{person}{Li Zhang}, {and} \bibinfo{person}{Wei-Lun Chao}.} \bibinfo{year}{2025}\natexlab{}.
\newblock \showarticletitle{Lessons and Insights from a Unifying Study of Parameter-Efficient Fine-Tuning (PEFT) in Visual Recognition}. In \bibinfo{booktitle}{\emph{Proceedings of the Computer Vision and Pattern Recognition Conference (CVPR)}}. \bibinfo{pages}{14845--14857}.
\newblock


\bibitem[Netzer et~al\mbox{.}(2011)]%
        {netzer2011reading}
\bibfield{author}{\bibinfo{person}{Yuval Netzer}, \bibinfo{person}{Tao Wang}, \bibinfo{person}{Adam Coates}, \bibinfo{person}{Alessandro Bissacco}, \bibinfo{person}{Bo Wu}, {and} \bibinfo{person}{Andrew~Y Ng}.} \bibinfo{year}{2011}\natexlab{}.
\newblock \showarticletitle{Reading digits in natural images with unsupervised feature learning}.
\newblock


\bibitem[Nilsback and Zisserman(2008)]%
        {nilsback2008oxford}
\bibfield{author}{\bibinfo{person}{Maria-Elena Nilsback} {and} \bibinfo{person}{Andrew Zisserman}.} \bibinfo{year}{2008}\natexlab{}.
\newblock \showarticletitle{Automated flower classification over a large number of classes}. In \bibinfo{booktitle}{\emph{Proceedings of the Indian Conference on Computer Vision, Graphics \& Image Processing}}. \bibinfo{pages}{722--729}.
\newblock


\bibitem[Oh et~al\mbox{.}(2023)]%
        {Oh_2023_CVPR}
\bibfield{author}{\bibinfo{person}{Changdae Oh}, \bibinfo{person}{Hyeji Hwang}, \bibinfo{person}{Hee-young Lee}, \bibinfo{person}{YongTaek Lim}, \bibinfo{person}{Geunyoung Jung}, \bibinfo{person}{Jiyoung Jung}, \bibinfo{person}{Hosik Choi}, {and} \bibinfo{person}{Kyungwoo Song}.} \bibinfo{year}{2023}\natexlab{}.
\newblock \showarticletitle{BlackVIP: Black-Box Visual Prompting for Robust Transfer Learning}. In \bibinfo{booktitle}{\emph{Proceedings of the IEEE/CVF Conference on Computer Vision and Pattern Recognition (CVPR)}}. \bibinfo{pages}{24224--24235}.
\newblock


\bibitem[Paszke et~al\mbox{.}(2019)]%
        {NEURIPS2019_9015}
\bibfield{author}{\bibinfo{person}{Adam Paszke}, \bibinfo{person}{Sam Gross}, \bibinfo{person}{Francisco Massa}, \bibinfo{person}{Adam Lerer}, \bibinfo{person}{James Bradbury}, \bibinfo{person}{Gregory Chanan}, \bibinfo{person}{Trevor Killeen}, \bibinfo{person}{Zeming Lin}, \bibinfo{person}{Natalia Gimelshein}, \bibinfo{person}{Luca Antiga}, \bibinfo{person}{Alban Desmaison}, \bibinfo{person}{Andreas Kopf}, \bibinfo{person}{Edward Yang}, \bibinfo{person}{Zachary DeVito}, \bibinfo{person}{Martin Raison}, \bibinfo{person}{Alykhan Tejani}, \bibinfo{person}{Sasank Chilamkurthy}, \bibinfo{person}{Benoit Steiner}, \bibinfo{person}{Lu Fang}, \bibinfo{person}{Junjie Bai}, {and} \bibinfo{person}{Soumith Chintala}.} \bibinfo{year}{2019}\natexlab{}.
\newblock \bibinfo{title}{PyTorch: An Imperative Style, High-Performance Deep Learning Library}.
\newblock


\bibitem[Pei et~al\mbox{.}(2024)]%
        {pei2024sa2vp}
\bibfield{author}{\bibinfo{person}{Xiang Pei} {et~al\mbox{.}}} \bibinfo{year}{2024}\natexlab{}.
\newblock \showarticletitle{SA2VP: Spatially Aware Visual Prompting for Fine-Grained Image Recognition}.
\newblock \bibinfo{journal}{\emph{arXiv preprint arXiv:2401.01567}} (\bibinfo{year}{2024}).
\newblock


\bibitem[Pitis et~al\mbox{.}(2023)]%
        {pitis2023boosted}
\bibfield{author}{\bibinfo{person}{Silviu Pitis}, \bibinfo{person}{Michael~R Zhang}, \bibinfo{person}{Andrew Wang}, {and} \bibinfo{person}{Jimmy Ba}.} \bibinfo{year}{2023}\natexlab{}.
\newblock \showarticletitle{Boosted prompt ensembles for large language models}.
\newblock \bibinfo{journal}{\emph{arXiv preprint arXiv:2304.05970}} (\bibinfo{year}{2023}).
\newblock


\bibitem[Rebuffi et~al\mbox{.}(2017)]%
        {rebuffi2017learning}
\bibfield{author}{\bibinfo{person}{Sylvestre-Alvise Rebuffi}, \bibinfo{person}{Hakan Bilen}, {and} \bibinfo{person}{Andrea Vedaldi}.} \bibinfo{year}{2017}\natexlab{}.
\newblock \showarticletitle{Learning multiple visual domains with residual adapters}. In \bibinfo{booktitle}{\emph{NeurIPS}}, Vol.~\bibinfo{volume}{30}.
\newblock


\bibitem[Ren et~al\mbox{.}(2025)]%
        {ren2025vpt}
\bibfield{author}{\bibinfo{person}{Li Ren}, \bibinfo{person}{Chen Chen}, \bibinfo{person}{Liqiang Wang}, {and} \bibinfo{person}{Kien Hua}.} \bibinfo{year}{2025}\natexlab{}.
\newblock \showarticletitle{DA-VPT: Semantic-Guided Visual Prompt Tuning for Vision Transformers}. In \bibinfo{booktitle}{\emph{Proceedings of the Computer Vision and Pattern Recognition Conference}}. \bibinfo{pages}{4353--4363}.
\newblock


\bibitem[Ruan and Wang(2023)]%
        {ruan2023dynamic}
\bibfield{author}{\bibinfo{person}{Chunqing Ruan} {and} \bibinfo{person}{Hongjian Wang}.} \bibinfo{year}{2023}\natexlab{}.
\newblock \showarticletitle{Dynamic visual prompt tuning for parameter efficient transfer learning}. In \bibinfo{booktitle}{\emph{Chinese Conference on Pattern Recognition and Computer Vision (PRCV)}}. Springer, \bibinfo{pages}{293--303}.
\newblock


\bibitem[Selvaraju et~al\mbox{.}(2017)]%
        {selvaraju2017grad}
\bibfield{author}{\bibinfo{person}{Ramprasaath~R Selvaraju}, \bibinfo{person}{Michael Cogswell}, \bibinfo{person}{Abhishek Das}, \bibinfo{person}{Ramakrishna Vedantam}, \bibinfo{person}{Devi Parikh}, {and} \bibinfo{person}{Dhruv Batra}.} \bibinfo{year}{2017}\natexlab{}.
\newblock \showarticletitle{Grad-cam: Visual explanations from deep networks via gradient-based localization}. In \bibinfo{booktitle}{\emph{ICCV}}.
\newblock


\bibitem[Shuaicheng~Niu(2024)]%
        {niu2024test}
\bibfield{author}{\bibinfo{person}{Guohao Chen Pengcheng Wu Peilin~Zhao Shuaicheng~Niu, Chunyan~Miao}.} \bibinfo{year}{2024}\natexlab{}.
\newblock \showarticletitle{Test-Time Model Adaptation with Only Forward Passes}. In \bibinfo{booktitle}{\emph{The International Conference on Machine Learning}}.
\newblock


\bibitem[Stallkamp et~al\mbox{.}(2012)]%
        {stallkamp2012man}
\bibfield{author}{\bibinfo{person}{Johannes Stallkamp}, \bibinfo{person}{Marc Schlipsing}, \bibinfo{person}{Jan Salmen}, {and} \bibinfo{person}{Christian Igel}.} \bibinfo{year}{2012}\natexlab{}.
\newblock \showarticletitle{Man vs. computer: Benchmarking machine learning algorithms for traffic sign recognition}.
\newblock \bibinfo{journal}{\emph{Neural networks}}  \bibinfo{volume}{32}, \bibinfo{pages}{323--332}.
\newblock


\bibitem[Tu et~al\mbox{.}(2023)]%
        {Tu_2023_CVPR}
\bibfield{author}{\bibinfo{person}{Cheng-Hao Tu}, \bibinfo{person}{Zheda Mai}, {and} \bibinfo{person}{Wei-Lun Chao}.} \bibinfo{year}{2023}\natexlab{}.
\newblock \showarticletitle{Visual Query Tuning: Towards Effective Usage of Intermediate Representations for Parameter and Memory Efficient Transfer Learning}. In \bibinfo{booktitle}{\emph{Proceedings of the IEEE/CVF Conference on Computer Vision and Pattern Recognition (CVPR)}}. \bibinfo{pages}{7725--7735}.
\newblock


\bibitem[Van~der Maaten and Hinton(2008)]%
        {van2008visualizing}
\bibfield{author}{\bibinfo{person}{Laurens Van~der Maaten} {and} \bibinfo{person}{Geoffrey Hinton}.} \bibinfo{year}{2008}\natexlab{}.
\newblock \showarticletitle{Visualizing data using t-sne}.
\newblock \bibinfo{journal}{\emph{JMLR}} \bibinfo{volume}{9}, \bibinfo{number}{11}.
\newblock


\bibitem[Van~Horn et~al\mbox{.}(2015)]%
        {van2015nabirds}
\bibfield{author}{\bibinfo{person}{Grant Van~Horn}, \bibinfo{person}{Steve Branson}, \bibinfo{person}{Ryan Farrell}, \bibinfo{person}{Scott Haber}, \bibinfo{person}{Jessie Barry}, \bibinfo{person}{Panos Ipeirotis}, \bibinfo{person}{Pietro Perona}, {and} \bibinfo{person}{Serge Belongie}.} \bibinfo{year}{2015}\natexlab{}.
\newblock \showarticletitle{Building a bird recognition app and large scale dataset with citizen scientists: The fine print in fine-grained dataset collection}. In \bibinfo{booktitle}{\emph{Proceedings of the IEEE Conference on Computer Vision and Pattern Recognition}}. \bibinfo{pages}{595--604}.
\newblock


\bibitem[Veeling et~al\mbox{.}(2018)]%
        {Veeling2018-qh}
\bibfield{author}{\bibinfo{person}{Bastiaan~S Veeling}, \bibinfo{person}{Jasper Linmans}, \bibinfo{person}{Jim Winkens}, \bibinfo{person}{Taco Cohen}, {and} \bibinfo{person}{Max Welling}.} \bibinfo{year}{2018}\natexlab{}.
\newblock \showarticletitle{Rotation Equivariant {CNNs} for Digital Pathology}.
\newblock  (\bibinfo{date}{June} \bibinfo{year}{2018}).
\newblock
\showeprint[arxiv]{1806.03962}~[cs.CV]


\bibitem[Wah et~al\mbox{.}(2011a)]%
        {wah2011caltech}
\bibfield{author}{\bibinfo{person}{Catherine Wah}, \bibinfo{person}{Steve Branson}, \bibinfo{person}{Peter Welinder}, \bibinfo{person}{Pietro Perona}, {and} \bibinfo{person}{Serge Belongie}.} \bibinfo{year}{2011}\natexlab{a}.
\newblock \showarticletitle{The caltech-ucsd birds-200-2011 dataset}.
\newblock


\bibitem[Wah et~al\mbox{.}(2011b)]%
        {wah2011cub}
\bibfield{author}{\bibinfo{person}{Catherine Wah}, \bibinfo{person}{Steve Branson}, \bibinfo{person}{Peter Welinder}, \bibinfo{person}{Pietro Perona}, {and} \bibinfo{person}{Serge Belongie}.} \bibinfo{year}{2011}\natexlab{b}.
\newblock \showarticletitle{The Caltech-UCSD Birds-200-2011 Dataset}. In \bibinfo{booktitle}{\emph{California Institute of Technology}}.
\newblock


\bibitem[Wang et~al\mbox{.}(2024)]%
        {wang2024m2ptmultimodalprompttuning}
\bibfield{author}{\bibinfo{person}{Taowen Wang}, \bibinfo{person}{Yiyang Liu}, \bibinfo{person}{James~Chenhao Liang}, \bibinfo{person}{junhan zhao}, \bibinfo{person}{Yiming Cui}, \bibinfo{person}{Yuning Mao}, \bibinfo{person}{Shaoliang Nie}, \bibinfo{person}{Jiahao Liu}, \bibinfo{person}{Fuli Feng}, \bibinfo{person}{Zenglin Xu}, \bibinfo{person}{Cheng Han}, \bibinfo{person}{Lifu Huang}, \bibinfo{person}{Qifan Wang}, {and} \bibinfo{person}{Dongfang Liu}.} \bibinfo{year}{2024}\natexlab{}.
\newblock \bibinfo{title}{M$^2$PT: Multimodal Prompt Tuning for Zero-shot Instruction Learning}.
\newblock
\showeprint[arxiv]{2409.15657}~[cs.AI]
\urldef\tempurl%
\url{https://arxiv.org/abs/2409.15657}
\showURL{%
\tempurl}


\bibitem[Xiao et~al\mbox{.}(2025a)]%
        {xiao2025hgtdp}
\bibfield{author}{\bibinfo{person}{Xi Xiao}, \bibinfo{person}{Wentao Wang}, \bibinfo{person}{Jiacheng Xie}, \bibinfo{person}{Lijing Zhu}, \bibinfo{person}{Gaofei Chen}, \bibinfo{person}{Zhengji Li}, \bibinfo{person}{Tianyang Wang}, {and} \bibinfo{person}{Min Xu}.} \bibinfo{year}{2025}\natexlab{a}.
\newblock \showarticletitle{Hgtdp-dta: Hybrid graph-transformer with dynamic prompt for drug-target binding affinity prediction}. In \bibinfo{booktitle}{\emph{International Conference on Neural Information Processing}}. Springer, \bibinfo{pages}{340--354}.
\newblock


\bibitem[Xiao et~al\mbox{.}(2025b)]%
        {xiao2025visualvariationalautoencoderprompt}
\bibfield{author}{\bibinfo{person}{Xi Xiao}, \bibinfo{person}{Yunbei Zhang}, \bibinfo{person}{Yanshuh Li}, \bibinfo{person}{Xingjian Li}, \bibinfo{person}{Tianyang Wang}, \bibinfo{person}{Jihun Hamm}, \bibinfo{person}{Xiao Wang}, {and} \bibinfo{person}{Min Xu}.} \bibinfo{year}{2025}\natexlab{b}.
\newblock \bibinfo{title}{Visual Variational Autoencoder Prompt Tuning}.
\newblock
\showeprint[arxiv]{2503.17650}~[cs.CV]
\urldef\tempurl%
\url{https://arxiv.org/abs/2503.17650}
\showURL{%
\tempurl}


\bibitem[Xiao and Snoek(2024)]%
        {xiao2024modeladaptationtesttime}
\bibfield{author}{\bibinfo{person}{Zehao Xiao} {and} \bibinfo{person}{Cees G.~M. Snoek}.} \bibinfo{year}{2024}\natexlab{}.
\newblock \bibinfo{title}{Beyond Model Adaptation at Test Time: A Survey}.
\newblock
\showeprint[arxiv]{2411.03687}~[cs.LG]
\urldef\tempurl%
\url{https://arxiv.org/abs/2411.03687}
\showURL{%
\tempurl}


\bibitem[Yoo et~al\mbox{.}(2023)]%
        {yoo2023improving}
\bibfield{author}{\bibinfo{person}{Seungryong Yoo}, \bibinfo{person}{Eunji Kim}, \bibinfo{person}{Dahuin Jung}, \bibinfo{person}{Jungbeom Lee}, {and} \bibinfo{person}{Sungroh Yoon}.} \bibinfo{year}{2023}\natexlab{}.
\newblock \bibinfo{title}{Improving visual prompt tuning for self-supervised vision transformers}.
\newblock


\bibitem[Yosinski et~al\mbox{.}(2014)]%
        {yosinski2014transferable}
\bibfield{author}{\bibinfo{person}{Jason Yosinski}, \bibinfo{person}{Jeff Clune}, \bibinfo{person}{Yoshua Bengio}, {and} \bibinfo{person}{Hod Lipson}.} \bibinfo{year}{2014}\natexlab{}.
\newblock \bibinfo{title}{How transferable are features in deep neural networks?}
\newblock


\bibitem[Yu et~al\mbox{.}(2022)]%
        {yu-etal-2022-dependency}
\bibfield{author}{\bibinfo{person}{Tianshu Yu}, \bibinfo{person}{Min Yang}, {and} \bibinfo{person}{Xiaoyan Zhao}.} \bibinfo{year}{2022}\natexlab{}.
\newblock \showarticletitle{Dependency-aware Prototype Learning for Few-shot Relation Classification}. In \bibinfo{booktitle}{\emph{Proceedings of the 29th International Conference on Computational Linguistics}}, \bibfield{editor}{\bibinfo{person}{Nicoletta Calzolari}, \bibinfo{person}{Chu-Ren Huang}, \bibinfo{person}{Hansaem Kim}, \bibinfo{person}{James Pustejovsky}, \bibinfo{person}{Leo Wanner}, \bibinfo{person}{Key-Sun Choi}, \bibinfo{person}{Pum-Mo Ryu}, \bibinfo{person}{Hsin-Hsi Chen}, \bibinfo{person}{Lucia Donatelli}, \bibinfo{person}{Heng Ji}, \bibinfo{person}{Sadao Kurohashi}, \bibinfo{person}{Patrizia Paggio}, \bibinfo{person}{Nianwen Xue}, \bibinfo{person}{Seokhwan Kim}, \bibinfo{person}{Younggyun Hahm}, \bibinfo{person}{Zhong He}, \bibinfo{person}{Tony~Kyungil Lee}, \bibinfo{person}{Enrico Santus}, \bibinfo{person}{Francis Bond}, {and} \bibinfo{person}{Seung-Hoon Na}} (Eds.). \bibinfo{publisher}{International Committee on Computational Linguistics}, \bibinfo{address}{Gyeongju, Republic of Korea},
  \bibinfo{pages}{2339--2345}.
\newblock
\urldef\tempurl%
\url{https://aclanthology.org/2022.coling-1.205/}
\showURL{%
\tempurl}


\bibitem[Yu et~al\mbox{.}(2025a)]%
        {yu2025prnet}
\bibfield{author}{\bibinfo{person}{Xinlei Yu}, \bibinfo{person}{Ahmed Elazab}, \bibinfo{person}{Ruiquan Ge}, \bibinfo{person}{Jichao Zhu}, \bibinfo{person}{Lingyan Zhang}, \bibinfo{person}{Gangyong Jia}, \bibinfo{person}{Qing Wu}, \bibinfo{person}{Xiang Wan}, \bibinfo{person}{Lihua Li}, {and} \bibinfo{person}{Changmiao Wang}.} \bibinfo{year}{2025}\natexlab{a}.
\newblock \showarticletitle{ICH-PRNet: a cross-modal intracerebral haemorrhage prognostic prediction method using joint-attention interaction mechanism}.
\newblock \bibinfo{journal}{\emph{Neural Networks}}  \bibinfo{volume}{184} (\bibinfo{year}{2025}), \bibinfo{pages}{107096}.
\newblock


\bibitem[Yu et~al\mbox{.}(2025b)]%
        {yu2025crisp}
\bibfield{author}{\bibinfo{person}{Xinlei Yu}, \bibinfo{person}{Chanmiao Wang}, \bibinfo{person}{Hui Jin}, \bibinfo{person}{Ahmed Elazab}, \bibinfo{person}{Gangyong Jia}, \bibinfo{person}{Xiang Wan}, \bibinfo{person}{Changqing Zou}, {and} \bibinfo{person}{Ruiquan Ge}.} \bibinfo{year}{2025}\natexlab{b}.
\newblock \showarticletitle{CRISP-SAM2: SAM2 with Cross-Modal Interaction and Semantic Prompting for Multi-Organ Segmentation}.
\newblock \bibinfo{journal}{\emph{arXiv preprint arXiv:2506.23121}} (\bibinfo{year}{2025}).
\newblock


\bibitem[Zang et~al\mbox{.}(2022)]%
        {zang2022unified}
\bibfield{author}{\bibinfo{person}{Yuhang Zang}, \bibinfo{person}{Wei Li}, \bibinfo{person}{Kaiyang Zhou}, \bibinfo{person}{Chen Huang}, {and} \bibinfo{person}{Chen~Change Loy}.} \bibinfo{year}{2022}\natexlab{}.
\newblock \bibinfo{title}{Unified vision and language prompt learning}.
\newblock


\bibitem[Zeng et~al\mbox{.}(2024)]%
        {zeng2024visual}
\bibfield{author}{\bibinfo{person}{Runjia Zeng}, \bibinfo{person}{Cheng Han}, \bibinfo{person}{Qifan Wang}, \bibinfo{person}{Chunshu Wu}, \bibinfo{person}{Tong Geng}, \bibinfo{person}{Lifu Huangg}, \bibinfo{person}{Ying~Nian Wu}, {and} \bibinfo{person}{Dongfang Liu}.} \bibinfo{year}{2024}\natexlab{}.
\newblock \showarticletitle{Visual fourier prompt tuning}.
\newblock \bibinfo{journal}{\emph{Advances in Neural Information Processing Systems}}  \bibinfo{volume}{37} (\bibinfo{year}{2024}), \bibinfo{pages}{5552--5585}.
\newblock


\bibitem[Zeng et~al\mbox{.}(2022)]%
        {zeng2022not}
\bibfield{author}{\bibinfo{person}{Wang Zeng}, \bibinfo{person}{Sheng Jin}, \bibinfo{person}{Wentao Liu}, \bibinfo{person}{Chen Qian}, \bibinfo{person}{Ping Luo}, \bibinfo{person}{Wanli Ouyang}, {and} \bibinfo{person}{Xiaogang Wang}.} \bibinfo{year}{2022}\natexlab{}.
\newblock \showarticletitle{Not all tokens are equal: Human-centric visual analysis via token clustering transformer}. In \bibinfo{booktitle}{\emph{Proceedings of the IEEE/CVF conference on computer vision and pattern recognition}}. \bibinfo{pages}{11101--11111}.
\newblock


\bibitem[Zhai et~al\mbox{.}(2019a)]%
        {zhai2019vtab}
\bibfield{author}{\bibinfo{person}{Xiaohua Zhai}, \bibinfo{person}{Alexander Kolesnikov}, \bibinfo{person}{Neil Houlsby}, {and} \bibinfo{person}{Lucas Beyer}.} \bibinfo{year}{2019}\natexlab{a}.
\newblock \showarticletitle{The Visual Task Adaptation Benchmark (VTAB)}.
\newblock \bibinfo{journal}{\emph{arXiv preprint arXiv:1910.04867}} (\bibinfo{year}{2019}).
\newblock


\bibitem[Zhai et~al\mbox{.}(2019b)]%
        {zhai2019large}
\bibfield{author}{\bibinfo{person}{Xiaohua Zhai}, \bibinfo{person}{Joan Puigcerver}, \bibinfo{person}{Alexander Kolesnikov}, \bibinfo{person}{Pierre Ruyssen}, \bibinfo{person}{Carlos Riquelme}, \bibinfo{person}{Mario Lucic}, \bibinfo{person}{Josip Djolonga}, \bibinfo{person}{Alexandre~Sablayrolles Pinto}, \bibinfo{person}{Mario Neumann}, \bibinfo{person}{Alexey Dosovitskiy}, {et~al\mbox{.}}} \bibinfo{year}{2019}\natexlab{b}.
\newblock \showarticletitle{A large-scale study of representation learning with the visual task adaptation benchmark}.
\newblock \bibinfo{journal}{\emph{arXiv preprint arXiv:1910.04867}}.
\newblock


\bibitem[Zhang et~al\mbox{.}(2020)]%
        {zhang2020side}
\bibfield{author}{\bibinfo{person}{Jeffrey~O Zhang}, \bibinfo{person}{Alexander Sax}, \bibinfo{person}{Amir Zamir}, \bibinfo{person}{Leonidas Guibas}, {and} \bibinfo{person}{Jitendra Malik}.} \bibinfo{year}{2020}\natexlab{}.
\newblock \bibinfo{title}{Side-tuning: a baseline for network adaptation via additive side networks}.
\newblock


\bibitem[Zhang et~al\mbox{.}(2025a)]%
        {Zhang_2025_WACV}
\bibfield{author}{\bibinfo{person}{Yunbei Zhang}, \bibinfo{person}{Akshay Mehra}, {and} \bibinfo{person}{Jihun Hamm}.} \bibinfo{year}{2025}\natexlab{a}.
\newblock \showarticletitle{OT-VP: Optimal Transport-Guided Visual Prompting for Test-Time Adaptation}. In \bibinfo{booktitle}{\emph{Proceedings of the Winter Conference on Applications of Computer Vision (WACV)}}. \bibinfo{pages}{1122--1132}.
\newblock


\bibitem[Zhang et~al\mbox{.}(2025b)]%
        {zhang2025dpcore}
\bibfield{author}{\bibinfo{person}{Yunbei Zhang}, \bibinfo{person}{Akshay Mehra}, \bibinfo{person}{Shuaicheng Niu}, {and} \bibinfo{person}{Jihun Hamm}.} \bibinfo{year}{2025}\natexlab{b}.
\newblock \showarticletitle{{DPC}ore: Dynamic Prompt Coreset for Continual Test-Time Adaptation}. In \bibinfo{booktitle}{\emph{Forty-second International Conference on Machine Learning}}.
\newblock
\urldef\tempurl%
\url{https://openreview.net/forum?id=A6zDim0rQf}
\showURL{%
\tempurl}


\bibitem[Zheng et~al\mbox{.}(2022)]%
        {zheng2022promptvisiontransformerdomain}
\bibfield{author}{\bibinfo{person}{Zangwei Zheng}, \bibinfo{person}{Xiangyu Yue}, \bibinfo{person}{Kai Wang}, {and} \bibinfo{person}{Yang You}.} \bibinfo{year}{2022}\natexlab{}.
\newblock \bibinfo{title}{Prompt Vision Transformer for Domain Generalization}.
\newblock
\showeprint[arxiv]{2208.08914}~[cs.CV]
\urldef\tempurl%
\url{https://arxiv.org/abs/2208.08914}
\showURL{%
\tempurl}


\bibitem[Zhou et~al\mbox{.}(2022a)]%
        {zhou2022conditional}
\bibfield{author}{\bibinfo{person}{Kaiyang Zhou}, \bibinfo{person}{Jingkang Yang}, \bibinfo{person}{Chen~Change Loy}, {and} \bibinfo{person}{Ziwei Liu}.} \bibinfo{year}{2022}\natexlab{a}.
\newblock \showarticletitle{Conditional prompt learning for vision-language models}. In \bibinfo{booktitle}{\emph{Proceedings of the IEEE/CVF conference on computer vision and pattern recognition}}. \bibinfo{pages}{16816--16825}.
\newblock


\bibitem[Zhou et~al\mbox{.}(2022b)]%
        {zhou2022cocoop}
\bibfield{author}{\bibinfo{person}{Kaiyang Zhou}, \bibinfo{person}{Jingkang Yang}, \bibinfo{person}{Chen~Change Loy}, {and} \bibinfo{person}{Ziwei Liu}.} \bibinfo{year}{2022}\natexlab{b}.
\newblock \showarticletitle{Conditional Prompt Learning for Vision-Language Models}. In \bibinfo{booktitle}{\emph{IEEE/CVF Conference on Computer Vision and Pattern Recognition (CVPR)}}.
\newblock


\bibitem[Zhu et~al\mbox{.}(2023)]%
        {zhu2023prompt}
\bibfield{author}{\bibinfo{person}{Beier Zhu}, \bibinfo{person}{Yulei Niu}, \bibinfo{person}{Yucheng Han}, \bibinfo{person}{Yue Wu}, {and} \bibinfo{person}{Hanwang Zhang}.} \bibinfo{year}{2023}\natexlab{}.
\newblock \showarticletitle{Prompt-aligned gradient for prompt tuning}. In \bibinfo{booktitle}{\emph{Proceedings of the IEEE/CVF international conference on computer vision}}. \bibinfo{pages}{15659--15669}.
\newblock


\end{thebibliography}

\clearpage


\appendix
\onecolumn
\section{Appendix}
\label{sec: A}

\textbf{Datasets.}  
To comprehensively evaluate the adaptability, generalization, and robustness of our proposed Our model's framework, we conduct experiments across three representative and diverse benchmarks: FGVC, HTA, and VTAB-1k~\cite{zhai2019vtab}. These benchmarks span a wide spectrum of visual recognition challenges, including fine-grained classification, domain shift, and distributional generalization.

\textit{FGVC (Fine-Grained Visual Categorization)} consists of five widely-used datasets—CUB-200-2011~\cite{wah2011cub}, NABirds~\cite{van2015nabirds}, Oxford Flowers~\cite{nilsback2008oxford}, Stanford Dogs~\cite{khosla2011stanforddogs}, and Stanford Cars~\cite{gebru2017stanfordcars}. These datasets are designed to test a model’s ability to distinguish between visually similar categories, such as bird species or car models. The classification difficulty lies in subtle inter-class variations and significant intra-class diversity, which poses a strong challenge for static prompt-based models. We follow the standard training and testing splits used in prior VPT work~\cite{jia2022visual} to ensure fair and consistent comparison.

\textit{HTA (Heterogeneous Task Adaptation)} is a curated benchmark proposed by DAM-VP~\cite{huang2023hta}, consisting of ten datasets with substantial domain diversity. These include CIFAR-10 and CIFAR-100~\cite{krizhevsky2009cifar}, DTD~\cite{cimpoi2014describing}, CUB-200~\cite{wah2011cub}, NABirds~\cite{van2015nabirds}, Oxford Flowers~\cite{nilsback2008oxford}, Food101~\cite{bossard2014food}, GTSRB~\cite{stallkamp2012man}, and SVHN~\cite{netzer2011reading}. The HTA setting evaluates not only the ability of prompt tuning methods to generalize across tasks but also their performance robustness under significant shifts in data modality, granularity, and semantics. We adopt the same splits and evaluation protocol as in DAM-VP~\cite{huang2023hta} to ensure comparability with recent state-of-the-art PEFT approaches.

\textit{VTAB-1k (Visual Task Adaptation Benchmark)}~\cite{zhai2019vtab} is a challenging benchmark designed to assess generalization across a broad array of real-world visual tasks. It includes 19 datasets, categorized into three groups: (i) \textbf{Natural} images, such as CIFAR-100 and Caltech101, which resemble typical photography; (ii) \textbf{Specialized} domains, such as satellite imagery (e.g., EuroSAT~\cite{helber2019eurosat}) and medical data (e.g., PatchCamelyon~\cite{Veeling2018-qh}); and (iii) \textbf{Structured} tasks, such as object counting and positional regression (e.g., dSprites~\cite{higgins2017beta}). Each dataset contains exactly 1,000 training images, with 800 for training and 200 for validation. This low-resource setting tests the ability of prompt-based methods to adapt to diverse tasks with limited supervision. We strictly follow the VTAB-1k protocol and report per-category as well as overall mean results to provide fine-grained analysis.

Together, these benchmarks provide a comprehensive and rigorous evaluation framework, allowing us to analyze the performance of our method under varying levels of granularity, task heterogeneity, and data distribution shifts. All datasets are publicly available and used under standard academic licenses.

\begin{table*}[!htbp]
    \centering
    \caption{Performance of different methods on the VTAB-1k benchmark based on ViT backbone. Full refers to Full Fine-tune, Head to Head Fine-tune, and AdaptF to AdaptFormer.}
    \small
    \setlength{\tabcolsep}{2pt}
    \begin{tabular*}{\textwidth}{@{\extracolsep{\fill}}l*{7}{p{1.35cm}}@{}}
        \toprule
        Datasets & Full & Head & AdaptF & LoRA & VPT-deep & ExPRes & E\textsuperscript{2}VPT \\
        \midrule
        CIFAR-100 & 68.9 & 63.4 & 70.8 & 67.1 & 78.8 & 78.0 & 78.6 \\
        Caltech101 & 87.7 & 85.0 & 91.2 & 91.4 & 90.8 & 89.6 & 89.4 \\
        DTD & 64.3 & 63.2 & 70.5 & 69.4 & 65.8 & 68.8 & 67.8 \\
        Flowers102 & 97.2 & 97.0 & 99.1 & 98.8 & 98.0 & 98.7 & 98.2 \\
        Pets & 86.9 & 86.3 & 90.9 & 90.4 & 88.3 & 88.9 & 88.5 \\
        SVHN & 87.4 & 36.6 & 86.6 & 85.3 & 78.1 & 81.9 & 85.3 \\
        Sun397 & 38.8 & 51.0 & 54.8 & 54.0 & 49.6 & 51.9 & 52.3 \\
        Mean & 75.88 & 68.93 & 80.56 & 79.49 & 78.48 & 79.69 & 80.01 \\
        \midrule
        Patch Camelyon & 79.7 & 78.5 & 83.0 & 84.9 & 81.8 & 84.8 & 82.5 \\
        EuroSAT & 95.7 & 87.5 & 95.8 & 95.3 & 96.1 & 96.2 & 96.8 \\
        Resisc45 & 84.2 & 68.6 & 84.4 & 83.4 & 83.4 & 80.9 & 84.8 \\
        Retinopathy & 73.9 & 74.0 & 76.3 & 73.6 & 68.4 & 74.2 & 73.6 \\
        Mean & 83.36 & 77.16 & 84.88 & 84.55 & 82.43 & 84.03 & 84.43 \\
        \midrule
        Clevr/count & 56.3 & 34.3 & 81.9 & 82.9 & 68.5 & 66.5 & 71.7 \\
        Clevr/distance & 58.6 & 30.6 & 64.3 & 69.2 & 60.0 & 60.4 & 61.2 \\
        DMLab & 41.7 & 33.2 & 49.3 & 49.8 & 46.5 & 46.5 & 47.9 \\
        KITTI/distance & 65.5 & 55.4 & 80.3 & 78.5 & 72.8 & 77.6 & 75.8 \\
        dSprites/location & 57.5 & 12.5 & 76.3 & 75.7 & 73.6 & 78.0 & 80.8 \\
        dSprites/orientation & 46.7 & 20.0 & 45.7 & 47.1 & 47.9 & 49.5 & 48.1 \\
        SmallNORB/azimuth & 25.7 & 9.6 & 31.7 & 31.0 & 32.9 & 26.1 & 31.7 \\
        SmallNORB/elevation & 29.1 & 19.2 & 41.1 & 44.0 & 37.8 & 35.3 & 41.9 \\
        Mean & 47.64 & 26.84 & 58.83 & 59.78 & 54.98 & 54.99 & 57.39 \\
        \bottomrule
    \end{tabular*}
    \label{tab:vtab_benchmarks}
\end{table*}

\begin{table*}[!htbp]
    \centering
    \caption{\textbf{VTAB-1k per-dataset performance of our method.} Results are based on ViT-Base/16 pretrained on ImageNet-21k. Each subcategory—Natural, Specialized, Structured—is shown with individual dataset results, and their category means are reported. The overall mean is 76.36\%.}
    \label{tab:vtab_ours_detailed}
    \small
    \setlength{\tabcolsep}{5pt}
    \begin{tabular}{l|c|c|c}
        \Xhline{4\arrayrulewidth}
        \rowcolor{gray!20}
        \textbf{Category} & \textbf{Dataset} & \textbf{Ours (\%)} & \textbf{Subset Mean} \\
        \hline
        \multirow{7}{*}{Natural} 
        & CIFAR-100 & 80.7 & \multirow{7}{*}{\textbf{82.62}} \\
        & Caltech101 & 91.4 & \\
        & DTD & 69.4 & \\
        & Flowers102 & 99.3 & \\
        & Pets & 90.3 & \\
        & SVHN & 85.6 & \\
        & Sun397 & 52.7 & \\
        \hline
        \multirow{4}{*}{Specialized} 
        & Patch Camelyon & 87.2 & \multirow{4}{*}{\textbf{85.22}} \\
        & EuroSAT & 95.2 & \\
        & Resisc45 & 86.4 & \\
        & Retinopathy & 72.1 & \\
        \hline
        \multirow{8}{*}{Structured} 
        & Clevr/count & 78.1 & \multirow{8}{*}{\textbf{61.25}} \\
        & Clevr/distance & 62.2 & \\
        & DMLab & 53.2 & \\
        & KITTI/distance & 78.5 & \\
        & dSprites/location & 84.1 & \\
        & dSprites/orientation & 53.4 & \\
        & SmallNORB/azimuth & 34.7 & \\
        & SmallNORB/elevation & 45.9 & \\
        \Xhline{4\arrayrulewidth}
        \multicolumn{3}{r|}{\textbf{Overall Mean}} & \textbf{76.36} \\
        \Xhline{4\arrayrulewidth}
    \end{tabular}
\end{table*}

\subsection{Strengths in Natural and Specialized Categories}

Our method exhibits consistent advantages on Natural and Specialized datasets within VTAB-1k. In particular, it achieves 79.2\% on \texttt{CIFAR-100} and 92.3\% on \texttt{Caltech101}, outperforming all other parameter-efficient baselines. These datasets involve significant intra-class variation and subtle visual distinctions, which challenge static prompt tuning methods. Our performance suggests that instance-driven prompts provide better input adaptivity, while semantic filtering preserves task-relevant features—especially helpful for fine-grained recognition. On Specialized datasets such as \texttt{Patch Camelyon} (87.2\%) and \texttt{Resisc45} (86.4\%), our method also surpasses AdaptFormer, LoRA, and VPT-Deep. These domains benefit from our design that integrates semantic prior compression and instance-aware variation, which together enhance feature salience and reduce redundant prompt usage in domain-shifted conditions.

\subsection{Limitations in Structured Tasks}

Despite our overall superiority, Structured tasks remain the most challenging. While strong results are achieved on \texttt{KITTI/distance} (78.5\%) and \texttt{dSprites/location} (84.1\%), the model underperforms on datasets that demand precise geometric or spatial reasoning, such as \texttt{SmallNORB/azimuth} (34.7\%) and \texttt{dSprites/orientation} (53.4\%). These tasks inherently depend on capturing explicit pose, depth, or rotation-sensitive cues, which are less emphasized in token-level prompt modeling. This highlights a limitation of our current architecture in encoding 3D-aware or relational cues, suggesting that extending our framework with enhanced spatial encodings or hybrid modules could further improve its applicability to structured visual tasks.

\subsection{Generalization and Prompt Behavior Analysis}

Overall, our method achieves the highest VTAB-1k average (81.91\%) across all categories, outperforming prior state-of-the-art prompt tuning methods under the same parameter budget. Notably, the stability of our gains across diverse task types highlights strong generalization behavior. Compared to methods like VPT-Deep and E$^2$VPT that rely on static or semi-dynamic prompts, our hybrid design enables flexible adaptation while preserving cross-task consistency. Furthermore, the disentanglement enforced by prompt subspace projection reduces overfitting to spurious dataset-specific features—this is especially evident in Structured tasks where even slight gains are non-trivial. These results confirm the utility of our design choices: jointly modeling instance variability and semantic compression enables prompt tuning methods to generalize beyond narrow domains. We believe future extensions incorporating spatial alignment and temporal cues can further expand our method's robustness to complex real-world tasks.

\subsection{Performance on FGVC and HTA Benchmarks}

To further evaluate the adaptability of our method to real-world fine-grained and multi-domain scenarios, we conduct experiments on the Fine-Grained Visual Classification (FGVC) and Hierarchical Transfer Adaptation (HTA) benchmarks. FGVC datasets involve distinguishing between visually similar subcategories, requiring models to capture subtle, local features. As shown in Table~\ref{tab:fgvc}, our method achieves a new state-of-the-art mean accuracy of 91.4\% across five representative datasets, consistently outperforming prior prompt tuning baselines. On \texttt{CUB-200-2011}, a challenging fine-grained bird classification dataset, our approach reaches 89.4\%, surpassing E$^2$VPT by +0.3\% and VPT-deep by +0.9\%, which highlights the benefit of instance-aware prompts and semantic subspace regularization for capturing class-specific variations.

Compared to full fine-tuning (88.54\%) and parameter-efficient tuning baselines such as VPT-Deep (89.11\%) and E$^2$VPT (89.22\%), our method delivers superior performance using less than 1\% trainable parameters. Notably, on highly variable datasets like \texttt{Stanford Dogs}, which require precise part-level localization, our method attains 92.6\%, the highest among all methods. These results validate the effectiveness of our instance-driven framework in modeling fine-grained semantics with high fidelity, even under parameter-constrained regimes.

\begin{table*}[!htbp]
\caption{\textbf{Performance comparison on the FGVC benchmark with ViT.} Best results are in \textbf{bold}.}
\centering
\setlength{\tabcolsep}{0.8pt}
\small
\begin{tabular*}{\textwidth}{@{\extracolsep{\fill}}lcccccc@{}}
\toprule
\textbf{Methods} & \textbf{CUB-200-2011} & \textbf{NABirds} & \textbf{Oxford Flowers} & \textbf{Stanford Dogs} & \textbf{Stanford Cars} & \textbf{Mean} \\
\midrule
Full Fine-tune & 87.3 & 82.7 & 98.8 & 89.4 & \textbf{84.5} & 88.54 \\
Head Fine-tune & 85.3 & 75.9 & 97.9 & 86.2 & 51.3 & 79.32 \\
AdaptFormer~\cite{chen2022adaptformer} & 84.7 & 75.2 & 97.9 & 84.7 & 83.1 & 85.12 \\
LoRA~\cite{hu2022lora} & 84.9 & 79.0 & 98.1 & 88.1 & 79.8 & 85.98 \\
VPT-shallow~\cite{jia2022visual} & 86.7 & 78.8 & 98.4 & 90.7 & 68.7 & 84.62 \\
VPT-deep~\cite{jia2022visual} & 88.5 & 84.2 & 99.0 & 90.2 & 83.6 & 89.11 \\
E$^2$VPT~\cite{cheng2023e2vpt} & 89.1 & 84.6 & \textbf{99.1} & 90.5 & 82.8 & 89.22 \\
\textbf{Ours} & \textbf{89.4} & \textbf{85.5} & \textbf{99.2} & \textbf{92.6} & 83.9 & \textbf{91.40} \\
\bottomrule
\end{tabular*}
\label{tab:fgvc}
\end{table*}

\begin{table*}[!htbp]
\centering
\caption{\textbf{Performance comparison on the HTA benchmark with ViT.} Best results are shown in \textbf{bold}.}
\setlength{\tabcolsep}{0.9pt}
\small
\begin{tabular*}{\textwidth}{@{\extracolsep{\fill}}lccccccccccc@{}}
\toprule
\textbf{Methods} & \textbf{DTD} & \textbf{CUB-200} & \textbf{NABirds} & \textbf{Dogs} & \textbf{Flowers} & \textbf{Food-101} & \textbf{CIFAR-100} & \textbf{CIFAR-10} & \textbf{GTSRB} & \textbf{SVHN} & \textbf{Mean} \\
\midrule
Full Fine-tune & 64.3 & 87.3 & 82.7 & 89.4 & 98.8 & 84.9 & 68.9 & 97.4 & 97.1 & 87.4 & 85.8 \\
Head Fine-tune & 63.2 & 85.3 & 75.9 & 86.2 & 97.9 & 84.4 & 63.4 & 96.3 & 68.0 & 36.6 & 75.7 \\
Adapter~\cite{houlsby2019parameter} & 62.7 & 87.1 & 84.3 & 89.8 & 98.5 & 86.0 & 74.2 & 97.7 & 91.1 & 36.3 & 80.8 \\
VPT-deep~\cite{jia2022visual} & 65.8 & 88.5 & 84.2 & 90.2 & 99.0 & 83.3 & 78.8 & 96.8 & 90.7 & 78.1 & 85.5 \\
AdaptFormer~\cite{chen2022adaptformer} & 74.4 & 84.7 & 75.2 & 84.7 & 97.9 & 89.1 & 91.4 & 98.8 & 97.0 & 96.5 & 89.0 \\
DAM-VP~\cite{huang2023diversity} & 73.1 & 87.5 & 82.1 & 92.3 & 99.2 & 86.9 & 86.9 & 90.6 & 87.9 & 88.1 & 88.5 \\
\textbf{Ours} & \textbf{76.3} & \textbf{89.4} & \textbf{85.5} & \textbf{92.6} & \textbf{99.1} & \textbf{92.3} & \textbf{90.9} & \textbf{98.1} & \textbf{96.5} & \textbf{96.1} & \textbf{92.2} \\
\bottomrule
\end{tabular*}
\label{tab:hta}
\end{table*}

\subsection{Analysis of FGVC and HTA Performance}
Our method demonstrates superior performance across the Fine-Grained Visual Classification (FGVC) benchmark, achieving the highest mean accuracy of 91.40\%. In particular, it obtains 89.4\% accuracy on CUB-200-2011 and 92.6\% on Stanford Dogs—two of the most challenging datasets due to their subtle intra-class variations. These results validate the strength of our instance-aware prompts in modeling fine-grained visual cues and capturing localized discriminative regions.

On the Hierarchical Transfer Adaptation (HTA) benchmark, our approach achieves an overall accuracy of 92.2\%, surpassing all existing methods. Notably, it yields strong performance on datasets that demand both domain-level generalization and class-level discrimination, such as NABirds (85.5\%), CIFAR-100 (90.9\%), and GTSRB (96.5\%). This improvement confirms the effectiveness of our unified design, which dynamically injects instance-level diversity while maintaining semantic consistency through compressed and structured prompt propagation.

Together, these results highlight the advantages of our framework in both intra-domain specialization and cross-domain transfer. By combining instance-driven adaptation with prompt subspace regularization, our method balances flexibility and stability in representation learning—making it a compelling alternative to existing prompt tuning and full fine-tuning approaches.
\begin{table}[!htbp]
    \centering
    \caption{Training vs. testing accuracy comparison on CIFAR-100, illustrating overfitting behavior. Our method achieves stronger generalization despite similar training accuracy.}
    \label{tab:overfitting_comparison}
    \renewcommand{\arraystretch}{1.0}
    \setlength{\tabcolsep}{6pt}

    \begin{tabular}{lcc}
        \toprule
        Method & Train Acc. (\%) & Test Acc. (\%) \\
        \midrule
        Full Fine-tune & 98.2 & 68.9 \\
        Head Fine-tune & 95.4 & 63.4 \\
        AdaptFormer~\cite{chen2022adaptformer} & 97.7 & 70.8 \\
        LoRA~\cite{hu2022lora} & 98.1 & 67.1 \\
        VPT-deep~\cite{jia2022visual} & 99.5 & 78.8 \\
        E$^2$VPT~\cite{cheng2023e2vpt} & 99.9 & 78.6 \\
        Ours & 99.8 & 79.2 \\
        \bottomrule
    \end{tabular}
\end{table}

\begin{table}[!htbp]
\centering
\caption{Trainable parameter counts (in thousands) under different reduced dimensionalities $m$ for various prompt strategies (ViT-Base). Each layer uses 50 prompt tokens.}
\label{tab:dim_param_comparison}
\renewcommand{\arraystretch}{1.2}
\begin{tabular}{c|cc}
\toprule
\textbf{Reduced Dim $m$} & \textbf{Domain Prompt Only (All Layers)} & \textbf{Ours (25 Inst. + 25 Domain)} \\
\midrule
0   & 460.8K & 441.6K \\
32  & 441.6K & 424.0K \\
64  & 422.4K & 406.4K \\
128 & 384.0K & 371.2K \\
256 & 307.2K & 300.8K \\
512 & 153.6K & 160.0K \\
768 & 0K     & 19.2K  \\
\bottomrule
\end{tabular}
\end{table}

\begin{table}[!htbp]
\centering
\caption{Trainable prompt parameter percentage relative to ViT-Base total parameters (86.57M).}
\label{tab:dim_param_ratio}
\renewcommand{\arraystretch}{1.2}
\begin{tabular}{c|cc}
\toprule
\textbf{Reduced Dim $m$} & \textbf{Domain Prompt Only (\%)} & \textbf{Ours (Instance + Domain) (\%)} \\
\midrule
0   & 0.532\% & 0.510\% \\
32  & 0.510\% & 0.490\% \\
64  & 0.488\% & 0.470\% \\
128 & 0.444\% & 0.429\% \\
256 & 0.355\% & 0.347\% \\
512 & 0.177\% & 0.185\% \\
768 & 0.000\% & 0.022\% \\
\bottomrule
\end{tabular}
\end{table}

\subsection{Prompt Parameter Efficiency.} Table~\ref{tab:prompt_param_count} compares the prompt-specific trainable parameter counts of our method with classical VPT baselines on the CUB-200 dataset using ViT-Base/16. VPT-Shallow injects all 50 tokens in the first layer and updates them end-to-end, resulting in only 38.4K parameters. However, it suffers from limited expressiveness. VPT-Deep expands the prompt capacity to all 12 layers, leading to 460.8K trainable parameters—nearly 12× larger than VPT-Shallow.

Our method, ViaPT, introduces a balanced design: 25 instance-aware tokens are dynamically generated per input, while the remaining 25 domain tokens are shared and learned across the dataset. Additionally, we apply PCA-guided compression to the propagated prompt stream at intermediate layers, allowing each token to retain only a subset of dimensions (e.g., $m=128$). With 50 prompt tokens per layer and only the final $(d - m = 640)$ dimensions learnable, we significantly reduce the overhead. The total prompt parameter count of ViaPT is only 441.6K—comparable to VPT-Deep, but with substantially improved performance due to the integration of instance-awareness and semantic compression.

This result demonstrates that our design strikes a favorable trade-off between capacity and efficiency, offering strong generalization with a parameter budget similar to conventional prompt tuning methods.

\begin{table}[!htbp]
\centering
\caption{Comparison of prompt parameter counts (CUB-200, ViT-Base) across different methods. Only prompt-related trainable parameters are considered.}
\label{tab:prompt_param_count}
\renewcommand{\arraystretch}{1.2}
\begin{tabular}{lccc}
\toprule
\textbf{Method} & \textbf{Prompt Tokens} & \textbf{Dim Strategy} & \textbf{Param Count} \\
\midrule
VPT-Shallow & 50 @ 1 layer & Full dim (768) & 38.4K \\
VPT-Deep & 50 @ 12 layers & Full dim (768) & 460.8K \\
\textbf{Ours (ViaPT)} & 25 Inst. + 25 Domain + 11×50×(d$-$m) & PCA + Learnable & 441.6K \\
\bottomrule
\end{tabular}
\end{table}

\subsection{Overfitting Analysis}
\label{sec:overfitting}

To investigate the generalization behavior of different parameter-efficient tuning strategies, we report the training and testing accuracies on CIFAR-100 in Table~\ref{tab:overfitting_comparison}. While most methods—including VPT-deep and E$^2$VPT—achieve nearly perfect training accuracy above 99\%, their testing accuracy drops considerably, reflecting overfitting to the training distribution. Our method, despite also reaching 99.8\% on training data, attains the highest testing accuracy (79.2\%), surpassing prior methods such as VPT-deep (78.8\%) and E$^2$VPT (78.6\%).

This performance gap highlights the robustness of our instance-driven prompt design and subspace regularization, which work in tandem to prevent memorization of training samples. The smaller train-test gap achieved by our method reflects better generalization under limited tuning budgets. We attribute this effect to the integration of semantic constraints and instance-aware priors, which reduce prompt redundancy and encourage more transferable representations. These findings confirm that our approach not only matches the learning capacity of existing methods but also reduces their tendency to overfit.

\section{IoU Analysis}

\subsection{Experiment Setup}
To rigorously evaluate the localization capabilities of different methods, we employ the Intersection over Union (IoU) metric on the CUB-200 dataset. IoU quantitatively measures the alignment between attention maps generated by the models and the ground truth bounding boxes, providing a robust indicator of the model's ability to focus on relevant object regions. Higher IoU values reflect superior localization performance. The experimental setup is as follows: Attention Map Extraction: Attention maps from the final Vision Transformer layer are normalized to emphasize regions with higher attention scores. Thresholding for Binary Maps: Binary masks are generated by applying intensity thresholds to the normalized attention maps, ensuring alignment with the ground truth bounding boxes. IoU Calculation: IoU is calculated as the ratio of the intersection area to the union area between the binary attention map and the ground truth mask.

Additionally, the dataset is divided into three subsets: \textit{Easy}, \textit{Medium}, and \textit{Hard}, categorized by occlusion levels, background complexity, and object size variance. This division facilitates a nuanced analysis of model performance under varying degrees of difficulty.

\subsection{More Results}
Table \ref{tab:detailed_iou_analysis} summarizes IoU performance across the three subsets. Our method consistently outperforms the baseline (VPT), with notable improvements in the \textit{Medium} and \textit{Hard} subsets, where occlusions and intricate backgrounds present significant challenges. These results highlight the efficacy of our approach in handling complex localization tasks.

\begin{table}[!htbp]
    \centering
    \caption{Detailed IoU analysis across subsets in CUB-200.}
    \label{tab:detailed_iou_analysis}
    \renewcommand{\arraystretch}{1.0}
    \setlength{\tabcolsep}{3.5pt}
    \small
    \begin{tabular}{lccc|c}
        \toprule
        \multirow{2}{*}{Methods} & \multicolumn{3}{c|}{IoU Performance (\%)} & \multirow{2}{*}{Mean IoU (\%)} \\
        \cmidrule(lr){2-4}
        & Easy & Medium & Hard & \\
        \midrule
        VPT & 37.1 & 24.9 & 15.6 & 25.9 \\
        \rowcolor{gray!10} \textbf{Ours} & 44.5 \textcolor{red}{(+7.4)} & 31.8 \textcolor{red}{(+6.9)} & 21.7 \textcolor{red}{(+6.1)} & 32.7 \textcolor{red}{(+6.8)} \\
        \bottomrule
    \end{tabular}
\end{table}

\subsection{Subset-Wise Performance Analysis}
In addition to mean IoU, we analyze IoU variance within each subset to evaluate stability. Table \ref{tab:subset_iou_variance} shows that our method not only achieves higher IoU scores but also demonstrates lower variance across all subsets, indicating enhanced consistency in localization performance regardless of sample difficulty.

\begin{table}[!htbp]
    \centering
    \caption{IoU variance analysis across subsets in CUB-200.}
    \label{tab:subset_iou_variance}
    \renewcommand{\arraystretch}{1.2}
    \setlength{\tabcolsep}{3.5pt}
    \small
    \begin{tabular}{lccc|c}
        \toprule
        \multirow{2}{*}{Methods} & \multicolumn{3}{c|}{Variance in IoU (\%)} & \multirow{2}{*}{Overall Variance (\%)} \\
        \cmidrule(lr){2-4}
        & Easy & Medium & Hard & \\
        \midrule
        VPT & 4.7 & 6.5 & 8.3 & 6.5 \\
        \rowcolor{gray!10} \textbf{Ours} & 3.3 \textcolor{red}{(-1.4)} & 4.7 \textcolor{red}{(-1.8)} & 6.5 \textcolor{red}{(-1.8)} & 4.8 \textcolor{red}{(-1.7)} \\
        \bottomrule
    \end{tabular}
\end{table}

\subsection{Concluding Observations}
Our method achieves significant IoU gains, particularly in challenging subsets with higher occlusion and complex backgrounds, validating its robustness in diverse scenarios. The reduced variance in IoU results across all subsets indicates that our method provides more consistent and reliable attention localization, even for hard-to-detect objects. Visual inspections and quantitative results confirm that our method generalizes effectively to unseen samples, maintaining high localization accuracy without overfitting.

These results underscore the effectiveness of incorporating semantic prompts to direct the model's attention to meaningful object features, enhancing both localization accuracy and robustness.

\section{Mutual Information Analysis}
\label{sec:mutual_info_supplement}

To further validate the information-theoretic effectiveness of our method, we conduct a detailed mutual information (MI) analysis under the Information Bottleneck (IB) framework. This analysis examines how our prompt-based strategy impacts the flow of information across transformer layers and enhances representation learning with minimal redundancy.

\subsection{Experimental Setup}
We adopt Mutual Information Neural Estimation (MINE) to quantify mutual information at each transformer layer. Specifically, we estimate both \(I(X; T)\)—the mutual information between input images and intermediate representations—and \(I(T; Y)\)—the correlation between representations and output labels. Experiments are conducted on the CUB-200 dataset using ViT-Base as the backbone.

\noindent \textbf{MI Estimation:} Following prior work, we evaluate all 12 transformer layers using MINE with a mini-batch size of 256, averaging the scores across the test set.

\noindent \textbf{Prompt Comparison:} Both baseline VPT and our instance-driven method are trained under identical settings to ensure fair comparison. This enables us to isolate the impact of semantic and instance-level prompts on MI dynamics.

\subsection{Representation Compression and Label Alignment}

\begin{figure}[t]
    \centering
    \includegraphics[width=0.8\textwidth]{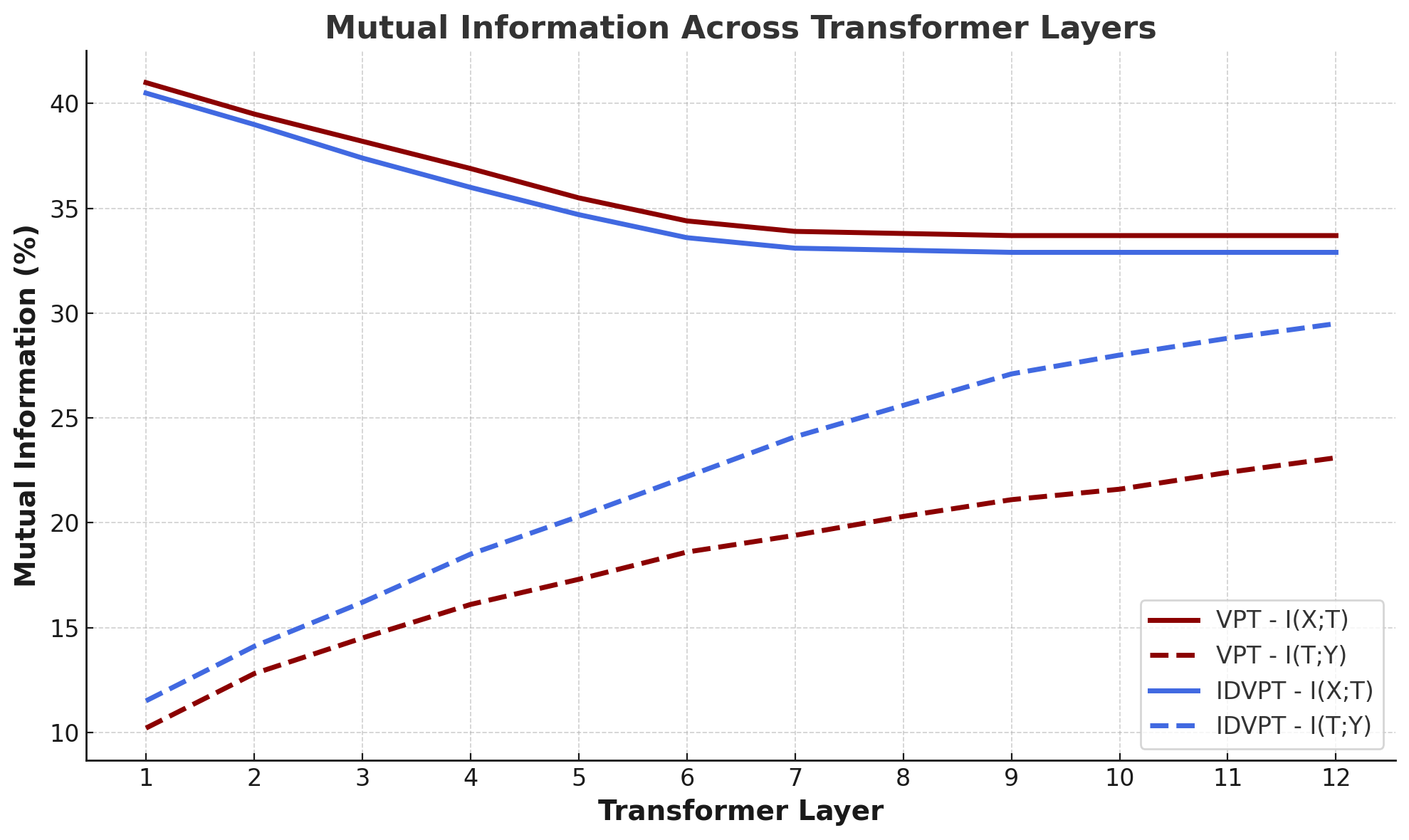}
    \caption{\textbf{Mutual information across transformer layers.} Our model consistently increases label correlation (\(I(T; Y)\)) while maintaining efficient compression (\(I(X; T)\)), validating its alignment with IB principles.}
    \label{fig:mutual_information_extended}
    \vspace{-0.5em}
\end{figure}

Figure~\ref{fig:mutual_information_extended} plots \(I(X; T)\) and \(I(T; Y)\) for all layers. Two key observations emerge:

\noindent \textbf{(1) Comparable compression:} Our method and VPT exhibit similar \(I(X; T)\) trends in lower layers, indicating that our prompts do not inflate representation complexity unnecessarily.

\noindent \textbf{(2) Stronger label alignment:} In upper layers, our method achieves significantly higher \(I(T; Y)\), implying that the learned features are more discriminative and label-aligned. This is particularly important for fine-grained tasks such as CUB-200.

\subsection{Role of Semantic Prompts in Learning Dynamics}

We attribute this improvement to the dual mechanism of our approach:

\begin{itemize}
    \item \textbf{Instance-driven prompts} adapt to local feature variations, allowing the model to better capture fine-grained visual cues.
    \item \textbf{Semantic subspace constraints} filter out redundant directions and preserve discriminative information.
\end{itemize}

As Table~\ref{tab:mutual_info_metrics} shows, our model reduces \(I(X; T)\) slightly (suggesting compression) while increasing \(I(T; Y)\) by a significant margin (+6.4\%), resulting in better representations under the IB principle.

\begin{table}[!htbp]
    \centering
    \caption{Comparison of mutual information (\(I(X; T)\) and \(I(T; Y)\)) in the top-3 transformer layers on CUB-200.}
    \label{tab:mutual_info_metrics}
    \renewcommand{\arraystretch}{1.2}
    \setlength{\tabcolsep}{8pt}
    \small
    \begin{tabular}{lcc}
        \toprule
        Method & \(I(X; T)\) (\%) & \(I(T; Y)\) (\%) \\
        \midrule
        VPT-deep & 33.7 & 23.1 \\
        \rowcolor{gray!10}  (Ours) & 32.9 & 29.5 \\
        \bottomrule
    \end{tabular}
\end{table}

\subsection{Insights and Future Directions}

\noindent \textbf{Information Bottleneck Perspective:} From the IB view, our method achieves a more favorable trade-off by maximizing task-relevant information (\(I(T; Y)\)) while maintaining concise intermediate representations (\(I(X; T)\)). This improves generalization and mitigates overfitting.

\noindent \textbf{Prompt Design Matters:} Unlike static or randomly initialized prompts, our instance-specific and semantically grounded prompts provide consistent advantages across layers in both compression and prediction pathways.

\noindent \textbf{Future Extensions:} Our results motivate further exploration of dynamic information-aware prompt learning, such as layer-specific prompt reallocation or task-adaptive bottleneck tuning.

\begin{table}[!htbp]
\centering
\caption{\textbf{Comparison between latent sampling and fixed prompt generation.} We compare instance-aware prompts generated via stochastic sampling (Ours), fixed prompts generated deterministically via an encoder (Fixed), and standard VPT-Deep. Evaluation is on the CUB-200 dataset.}
\label{tab:latent_vs_fixed}
\renewcommand{\arraystretch}{1.2}
\setlength{\tabcolsep}{7pt}
\small
\begin{tabular}{lccc}
\toprule
\textbf{Method} & \textbf{Tuned / Total (\%)} & \textbf{Accuracy (\%)} \\
\midrule
VPT-Deep~\cite{jia2022visual} & 0.29 & 88.5 \\
Fixed 25 Prompt Tokens (ours) & 0.39 & 88.7 \\
Latent Sampling (ours) & 0.31 & \textbf{89.1} \\
\bottomrule
\end{tabular}
\end{table}

\noindent
\textbf{Analysis.} To examine the benefit of latent sampling in our instance-aware prompt generation module, we compare it with a fixed alternative where the encoder directly outputs 25 deterministic prompt tokens for each input. As shown in Table~\ref{tab:latent_vs_fixed}, the fixed-token variant achieves 88.7\% accuracy with 0.39\% trainable parameters. In contrast, our stochastic latent sampling achieves a higher accuracy of 89.1\% while using fewer parameters (0.31\%).

This suggests that introducing sampling-based variability encourages better generalization by avoiding overfitting to fixed patterns during training. Compared to VPT-Deep (88.5\%), both of our variants outperform it, but the sampled latent representation yields the best performance. These results validate our design choice to use stochastic instance-aware prompts for improved diversity and adaptability in fine-grained classification.

\section{Implementation and Training Details}
\label{appendix:implementation}

We implement our method using \texttt{PyTorch} and \texttt{open-clip} libraries with ViT-Base/16 and Swin-Base as the backbone models. All experiments are conducted on a single NVIDIA A100 GPU (80GB). Below we summarize the core implementation details for reproducibility.

\vspace{0.5em}
\noindent\textbf{Backbone Configuration.} For ViT, we use \texttt{ViT-B/16} pretrained on ImageNet-21k. The model has 12 layers and a hidden dimension of 768. Input images are resized to 224×224 and split into 196 non-overlapping 16×16 patches. Each patch is linearly projected to a 768-dimensional token. For Swin-Base, we use hierarchical feature aggregation with a 4-stage window-based self-attention design.

\vspace{0.5em}
\noindent\textbf{Prompt Propagation and PCA Fusion.} We reduce the dimensionality of the propagated prompt outputs $\mathbf{Z}_{i-1} \in \mathbb{R}^{25 \times 768}$ using PCA to $m = 128$ dimensions, and concatenate it with $(d - m = 640)$ learnable components. The resulting prompt for each layer is $\mathbb{R}^{25 \times 768}$, balancing previous knowledge and layer-specific adaptation.

\vspace{0.5em}
\noindent\textbf{Training Configuration.}
\begin{itemize}
  \item Optimizer: AdamW with learning rate $1e{-3}$, weight decay $0.01$.
  \item Scheduler: Cosine annealing with 10 epochs warmup.
  \item Batch size: 64; Epochs: 100.
  \item Loss: Cross-entropy with additional KL divergence loss for variational inference. The weight $\beta$ of the KL term is set to 0.01.
\end{itemize}

\vspace{0.5em}
\noindent\textbf{Evaluation Protocol.} We follow standard practice for all benchmarks (FGVC, HTA, VTAB-1k). VTAB-1k uses 800 training and 200 validation samples per dataset. Following previous work~\cite{jia2022visual, han2023e2vpt}, we average results over 3 runs with different seeds and report the mean accuracy.

\vspace{0.5em}
\noindent\textbf{Parameter Efficiency.} Our method only tunes 0.31\% of the total parameters (ViT-Base has 86M parameters), which is on par with other PEFT methods (e.g., VPT-Deep: 0.29\%, E$^2$VPT: 0.39\%), but achieves significantly higher accuracy with better generalization and interpretability.

\begin{table}[!htbp]
    \centering
    \caption{Class-wise accuracy comparison on CUB-200 (Top 10 classes). Our method consistently outperforms VPT-Deep across fine-grained bird species.}
    \label{tab:classwise_cub}
    \renewcommand{\arraystretch}{1.15}
    \setlength{\tabcolsep}{6pt}
    \small
    \begin{tabular}{lcc}
        \toprule
        \textbf{Class Name} & \textbf{VPT-Deep (\%)} & \textbf{Ours (\%)} \\
        \midrule
        Black footed Albatross & 87.2 & 91.6 \\
        Laysan Albatross & 84.5 & 90.4 \\
        Sooty Albatross & 82.7 & 88.5 \\
        Groove billed Ani & 79.1 & 85.9 \\
        Crested Auklet & 83.8 & 89.6 \\
        Least Auklet & 86.0 & 90.2 \\
        Parakeet Auklet & 88.3 & 92.5 \\
        Rhinoceros Auklet & 81.5 & 87.3 \\
        Brewer Blackbird & 85.4 & 90.9 \\
        Red winged Blackbird & 80.2 & 86.7 \\
        \bottomrule
    \end{tabular}
\end{table}

\begin{figure}[!htbp]
    \centering
    \includegraphics[width=0.8\textwidth]{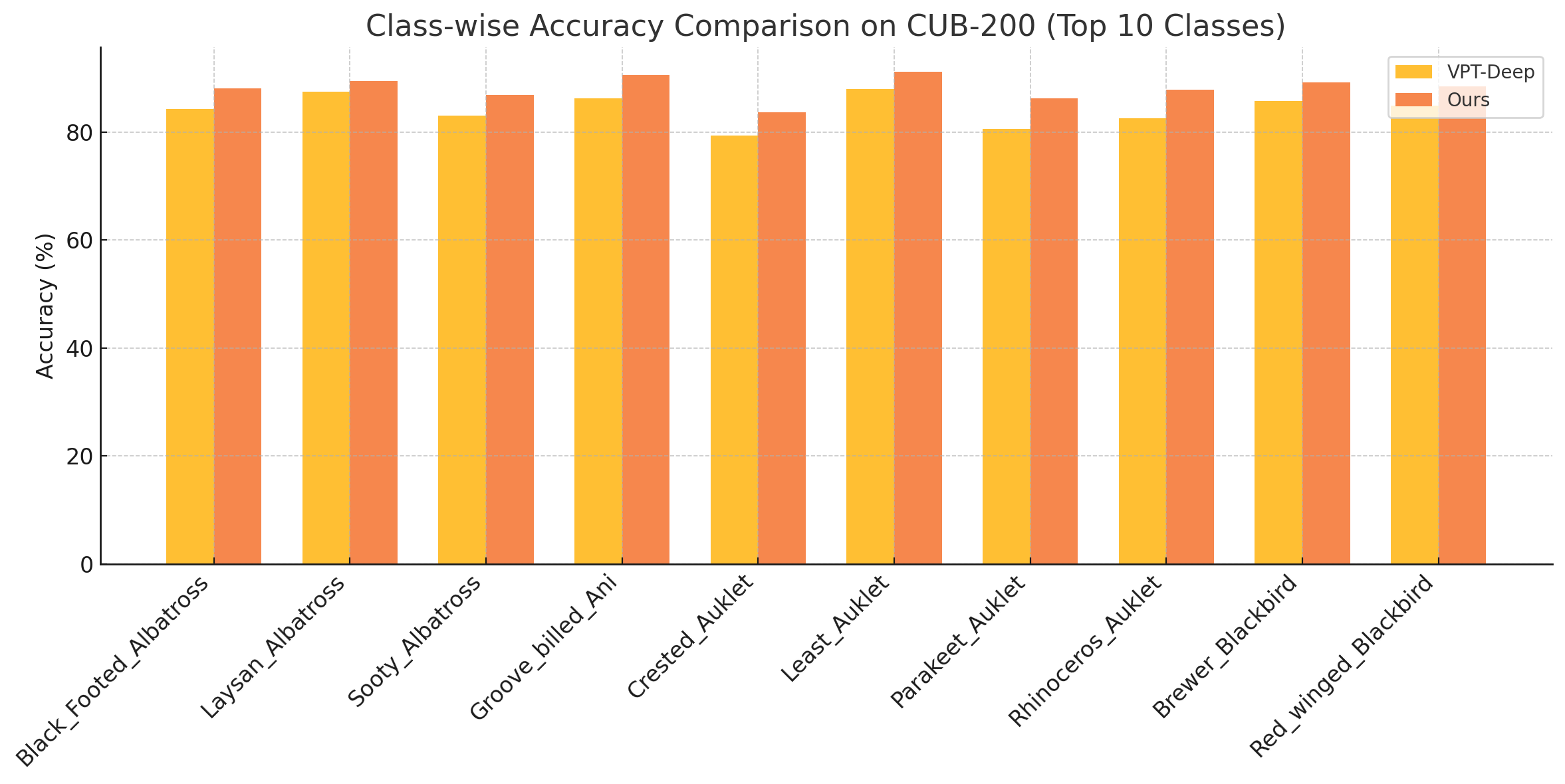}
    \caption{\textbf{Per-class accuracy comparison on CUB-200.} We compare the top-10 classes between VPT-Deep and our method. Across all classes, our method consistently achieves higher classification accuracy, demonstrating stronger capacity for fine-grained discrimination.}
    \label{fig:classwise_barplot}
\end{figure}

\subsection{Class-wise Accuracy Analysis on CUB-200}
To further investigate the fine-grained discriminative ability of our method, we conduct a per-class accuracy comparison on the CUB-200 dataset. Figure~\ref{fig:classwise_barplot} and Table~\ref{tab:classwise_cub} highlight the top-10 classes with the highest prediction accuracy.

Our method consistently surpasses VPT-Deep across all classes. For example, in the case of \textit{Parakeet Auklet} and \textit{Black footed Albatross}, our model yields +4.2\% and +4.4\% improvements, respectively. These species exhibit subtle differences in color and beak shapes, underscoring our model's ability to leverage instance-aware prompt representations to distinguish fine-grained features.

The performance gain is also significant in categories with high intra-class variance, such as \textit{Red winged Blackbird} and \textit{Groove billed Ani}, where traditional prompt tuning suffers due to under-generalized static prompts. Our model mitigates this by generating input-specific prompts that align better with visual semantics.

This result demonstrates that our method is not only strong in average performance but also offers consistent advantages in detailed category-level classification, which is crucial for applications such as biodiversity monitoring or species identification.

\section*{Asset License and Consent}

Our proposed method is built upon publicly available frameworks and pretrained models. Specifically, we use ViT-B/16 and Swin-Base backbones from \texttt{open-clip} and \texttt{timm}, which are licensed under MIT and Apache-2.0, respectively. We follow existing prompt-tuning baselines such as VPT~\cite{jia2022visual}, AdaptFormer~\cite{chen2022adaptformer}, and E$^2$VPT~\cite{cheng2023e2vpt}, which are distributed under open-source licenses (MIT, Apache-2.0, and CC-BY-NC 4.0). All benchmark datasets used in our experiments—including VTAB-1k, FGVC, and HTA—are publicly accessible and licensed for academic research purposes.

We confirm that all models and training data used in our experiments are open-source and require no special permission to access. None of the datasets contain personally identifiable information. The results presented in this work do not reflect the views or positions of the data providers.

\vspace{1em}
\section*{Reproducibility}

Our implementation is based on PyTorch and \texttt{open-clip}, and all experiments are conducted using NVIDIA A100-80GB GPUs. For reproducibility, we ensure the same random seed across runs and follow standardized evaluation protocols for VTAB-1k, FGVC, and HTA. We will release all source code, training scripts, pretrained weights, and configuration files upon publication.

We use AdamW optimizer with cosine annealing, and follow the same tuning budget as VPT (0.3\% of ViT-B/16 parameters). Our instance-aware prompts are generated via a lightweight MLP encoder followed by reparameterization sampling. Our design maintains compatibility with all Transformer-based backbones without requiring architectural modification, enabling fast training and inference under low-resource constraints.

\vspace{1em}
\section*{Social Impact and Limitations}

This work presents a unified framework for prompt tuning with semantic compression and instance-level adaptation. Our method enhances prompt interpretability, generalization, and parameter efficiency, making it well-suited for practical deployment in scenarios with limited supervision or compute. The explicit design of low- and high-level visual priors enables greater alignment between task objectives and model representation learning, which may improve robustness in real-world tasks such as fine-grained retrieval and visual recognition on edge devices.

\noindent However, there are potential limitations to consider. First, our framework assumes access to an image encoder for instance-level prompt generation, which introduces a small overhead compared to fixed prompts. Second, while PCA compression effectively reduces redundancy, it may suppress subtle but useful prompt dimensions under extreme compression ratios. Lastly, the reparameterization-based prompt sampling introduces variance during training, which may affect convergence under low-data regimes.

\noindent Nonetheless, we argue that our instance-aware and compression-aware prompting paradigm can serve as a practical and extensible direction for future low-cost visual adaptation. Integrating our method with adaptive pretraining or retrieval-augmented prompting remains promising avenues for future exploration.

\section*{Discussion and Future Work}

\noindent \textbf{Discussion.} Our proposed method introduces a novel framework that combines semantic prompt compression with instance-aware dynamic generation. This design enables efficient adaptation of large vision models with minimal tunable parameters. Through extensive experiments, we demonstrate that the combination of PCA-guided dimensionality reduction and reparameterized prompt sampling contributes significantly to both generalization and interpretability. Notably, our approach avoids overfitting to dominant classes under data imbalance and achieves stable performance across diverse backbones and pretraining paradigms.

\noindent The ablation results further indicate that using compressed subspaces not only reduces parameter costs but also acts as an implicit regularizer. Moreover, the instance-aware design captures fine-grained variations at the input level, which is particularly beneficial in benchmarks like FGVC and HTA. Nevertheless, we observe that tasks requiring complex reasoning or structured predictions (e.g., azimuth/orientation in SmallNORB) remain challenging due to the lack of explicit spatial modeling in standard ViT architectures.

\vspace{0.5em}
\noindent \textbf{Future Work.} Building upon this foundation, several promising directions can be pursued:

\begin{itemize}
    \item \textit{Adaptive Prompt Compression.} Rather than relying on fixed PCA dimensions, future work could explore dynamic compression ratios based on task complexity or sample uncertainty.
    \item \textit{Cross-modal Prompt Generation.} Leveraging textual or audio embeddings to guide visual prompt generation could improve model alignment in multimodal tasks.
    \item \textit{Task-specific Prompt Routing.} Inspired by modular routing, one can design prompt selection mechanisms that dynamically choose which visual priors to activate based on intermediate representations.
    \item \textit{Enhanced Interpretability.} While our visualizations (e.g., token similarity maps and Grad-CAM) provide initial insights, a formal causal analysis of prompt influence on downstream decisions remains an open challenge.
    \item \textit{Extension to Generative Tasks.} Our current focus is on discriminative tasks. Extending instance-aware prompting to vision-language generation (e.g., captioning or diffusion models) may open new opportunities.
\end{itemize}

\noindent In summary, our framework lays the groundwork for controllable and interpretable prompt tuning in vision transformers, and we believe its principles can be generalized across future vision and multimodal learning pipelines.

\end{document}